\documentclass{article}

\usepackage[nonatbib,final]{nips_2017}

\usepackage[utf8]{inputenc} 
\usepackage[T1]{fontenc}    
\usepackage{hyperref}       
\usepackage{url}            
\usepackage{booktabs}       
\usepackage{amsfonts}       
\usepackage{amsmath}
\usepackage{amsthm}
\usepackage{amssymb}
\usepackage{nicefrac}       
\usepackage{microtype}      
\usepackage{graphicx}
\usepackage{caption}
\usepackage{subcaption}

\newtheorem{theorem}{Theorem}

\title{Safe and Nested Subgame Solving for Imperfect-Information Games\thanks{A version of this paper was posted on the authors’ web pages in 2016, submitted to the AAAI-17 Workshop on Computer Poker and Imperfect Information Games in October 2016, and published in that workshop on February 5th, 2017.}
}

\author{
	Noam Brown \\
	Computer Science Department\\
	Carnegie Mellon University\\
	Pittsburgh, PA 15217 \\
	\texttt{noamb@cs.cmu.edu} \\
	\And
	Tuomas Sandholm \\
	Computer Science Department \\
	Carnegie Mellon University \\
	Pittsburgh, PA 15217 \\
	\texttt{sandholm@cs.cmu.edu}
}

\begin{document}	
	\maketitle
	\begin{abstract}
		In imperfect-information games, the optimal strategy in a subgame may depend on the strategy in other, unreached subgames. Thus a subgame cannot be solved in isolation and must instead consider the strategy for the entire game as a whole, unlike perfect-information games. Nevertheless, it is possible to first approximate a solution for the whole game and then improve it by solving individual subgames.
		This is referred to as \emph{subgame solving}.
		We introduce subgame-solving techniques that outperform prior methods both in theory and practice. We also show how to adapt them, and past subgame-solving techniques, to respond to opponent actions that are outside the original action abstraction; this significantly outperforms the prior state-of-the-art approach, action translation. Finally, we show that subgame solving can be repeated as the game progresses down the game tree, leading to far lower exploitability. These techniques were a key component of \emph{Libratus}, the first AI to defeat top humans in heads-up no-limit Texas hold'em poker.
	\end{abstract}
	
	\section{Introduction}
	\label{sec:intro}
	
	Imperfect-information games model strategic settings that have hidden information. They have a myriad of applications including negotiation, auctions, cybersecurity, and physical security.
	
	In perfect-information games, determining the optimal strategy at a decision point only requires knowledge of the game tree's current node and the remaining game tree beyond that node (the \emph{subgame} rooted at that node). This fact has been leveraged by nearly every AI for perfect-information games, including AIs that defeated top humans in chess~\cite{Campbell02:Deep} and Go~\cite{Silver16:Mastering}.
	In checkers, the ability to decompose the game into smaller independent subgames was even used to solve the entire game~\cite{Schaeffer07:Checkers}.
	However, it is not possible to determine a subgame's optimal strategy in an imperfect-information game using only knowledge of that subgame, because the game tree's exact node is typically unknown. Instead, the optimal strategy may depend on the value an opponent could have received in some other, unreached subgame. Although this is counter-intuitive, we provide a demonstration in Section~\ref{sec:cointoss}.
	
	Rather than rely on subgame decomposition, past approaches for imperfect-information games typically solved the game as a whole upfront. For example, heads-up limit Texas hold'em, a relatively simple form of poker with $10^{13}$ decision points, was essentially solved without decomposition~\cite{Bowling15:Heads-up}. However, this approach cannot extend to larger games, such as heads-up no-limit Texas hold'em---the primary benchmark in imperfect-information game solving---which has $10^{161}$ decision points~\cite{Johanson13:Measuring}.
	
	The standard approach to computing strategies in such large games is to first generate an \emph{abstraction} of the game, which is a smaller version of the game that retains as much as possible the strategic characteristics of the original game~\cite{Sandholm10:State,Sandholm15:Solving,Sandholm15:Abstraction}. For example, a continuous action space might be discretized. This abstract game is solved and its solution is used when playing the full game by mapping states in the full game to states in the abstract game. We refer to the solution of an abstraction (or more generally any approximate solution to a game) as a \emph{blueprint} strategy.
	
	In heavily abstracted games, a blueprint may be far from the true solution.
	\emph{Subgame solving} attempts to improve upon the blueprint by solving in real time a more fine-grained abstraction for an encountered subgame, while fitting its solution within the overarching blueprint.
	
	\section{Coin Toss}
	\label{sec:cointoss}
\begin{figure}[!h]
		\centering
		\includegraphics[width=100mm]{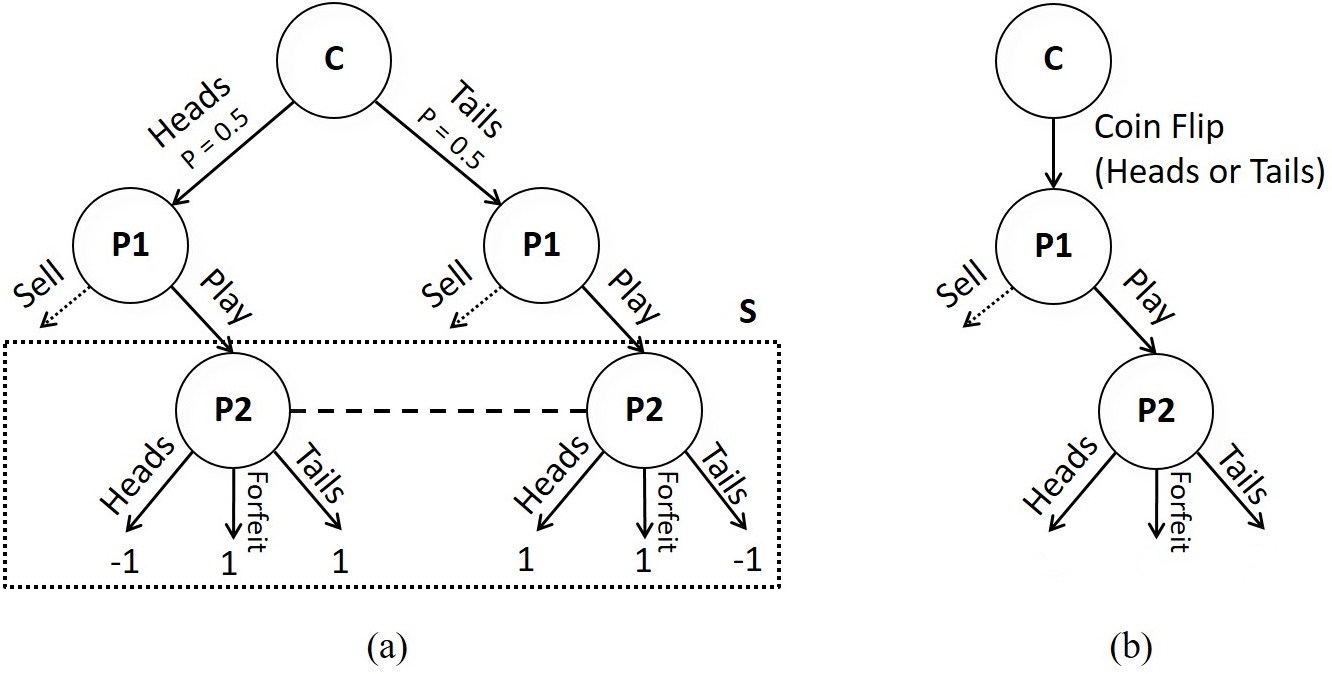}
		\caption{\small{(a) The example game of Coin Toss. ``C'' represents a chance node. $S$ is a Player 2 ($P_2$) subgame. The dotted line between the two $P_2$ nodes means that $P_2$ cannot distinguish between them. (b) The public game tree of Coin Toss. The two outcomes of the coin flip are only observed by $P_1$.}}
		\label{fig:cointoss}
	\end{figure}

	In this section we provide intuition for why an imperfect-information subgame cannot be solved in isolation. We demonstrate this in a simple game we call Coin Toss, shown in Figure~\ref{fig:cointoss}a, which will be used as a running example throughout the paper.
	
	Coin Toss is played between players $P_1$ and $P_2$. The figure shows rewards only for $P_1$; $P_2$ always receives the negation of $P_1$'s reward. A coin is flipped and lands either Heads or Tails with equal probability, but only $P_1$ sees the outcome. $P_1$ then chooses between actions ``Sell'' and ``Play.'' The Sell action leads to a subgame whose details are not important, but the \emph{expected value} (EV) of choosing the Sell action will be important.
	(For simplicity, one can equivalently assume \emph{in this section} that Sell leads to an immediate terminal reward, where the value depends on whether the coin landed Heads or Tails).
	If the coin lands Heads, it is considered lucky and $P_1$ receives an EV of $\$0.50$ for choosing Sell. On the other hand, if the coin lands Tails, it is considered unlucky and $P_1$ receives an EV of $-\$0.50$ for action Sell.
	(That is, $P_1$ must on average pay $\$0.50$ to get rid of the coin).
	If $P_1$ instead chooses Play, then $P_2$ may guess how the coin landed. If $P_2$ guesses correctly, then $P_1$ receives a reward of $-\$1$. If $P_2$ guesses incorrectly, then $P_1$ receives $\$1$. $P_2$ may also forfeit, which should never be chosen but will be relevant in later sections. We wish to determine the optimal strategy for $P_2$ in the subgame $S$ that occurs after $P_1$ chooses Play, shown in Figure~\ref{fig:cointoss}a.
	
	Were $P_2$ to always guess Heads, $P_1$ would receive $\$0.50$ for choosing Sell when the coin lands Heads, and $\$1$ for Play when it lands Tails. This would result in an average of $\$0.75$ for $P_1$.
	Alternatively, were $P_2$ to always guess Tails, $P_1$ would receive $\$1$ for choosing Play when the coin lands Heads, and $-\$0.50$ for choosing Sell when it lands Tails. This would result in an average reward of $\$0.25$ for $P_1$.
	However, $P_2$ would do even better by guessing Heads with $25\%$ probability and Tails with $75\%$ probability. In that case, $P_1$ could only receive $\$0.50$ (on average) by choosing Play when the coin lands Heads---the same value received for choosing Sell. Similarly, $P_1$ could only receive $-\$0.50$ by choosing Play when the coin lands Tails, which is the same value received for choosing Sell. This would yield an average reward of $\$0$ for $P_1$. It is easy to see that this is the best $P_2$ can do, because $P_1$ can average $\$0$ by always choosing Sell. Therefore, choosing Heads with $25\%$ probability and Tails with $75\%$ probability is an optimal strategy for $P_2$ in the ``Play'' subgame.
	
	Now suppose the coin is considered lucky if it lands Tails and unlucky if it lands Heads. That is, the expected reward for selling the coin when it lands Heads is now $-\$0.50$ and when it lands Tails is now $\$0.50$. It is easy to see that $P_2$'s optimal strategy for the ``Play'' subgame is now to guess Heads with $75\%$ probability and Tails with $25\%$ probability. This shows that a player's optimal strategy in a subgame can depend on the strategies and outcomes in other parts of the game. Thus, one cannot solve a subgame using information about that subgame alone. This is the central challenge of imperfect-information games as opposed to perfect-information games. 
	
	\section{Notation and Background}
	\label{sec:background}
	This paper focuses on two-player zero-sum games.
	In a two-player zero-sum extensive-form game there are two players, $\mathcal{P} = \{1,2\}$. $H$ is the set of all possible nodes, represented as a sequence of actions. $A(h)$ is the actions available in a node and $P(h) \in \mathcal{P} \cup c$ is the player who acts at that node, where $c$ denotes chance. Chance plays an action $a \in A(h)$ with a fixed probability. If action $a \in A(h)$ leads from $h$ to $h'$, then we write $h \cdot a = h'$. If a sequence of actions leads from $h$ to $h'$, then we write $h \sqsubset h'$. The set of nodes $Z \subseteq H$ are terminal nodes. For each player $i \in \mathcal{P}$, there is a payoff function $u_i: Z\rightarrow \Re$ where $u_1 = -u_2$.
	
	Imperfect information is represented by \emph{information sets} (infosets). Every node $h \in H$ belongs to exactly one infoset for each player. For any infoset $I_i$, nodes $h, h' \in I_i$ are indistinguishable to player~$i$. Thus the same player must act at all the nodes in an infoset, and the same actions must be available.
	Let $P(I_i)$ and $A(I_i)$ be such that all $h \in I_i$, $P(I_i) = P(h)$ and $A(I_i) = A(h)$.
	
	A strategy $\sigma_i(I_i)$ is a probability vector over $A(I_i)$ for infosets where $P(I_i) = i$. The probability of action $a$ is denoted by $\sigma_i(I_i,a)$. For all $h \in I_i$, $\sigma_i(h) = \sigma_i(I_i)$.
	A full-game strategy $\sigma_i \in \Sigma_i$ defines a strategy for each player~$i$ infoset. A strategy profile $\sigma$ is a tuple of strategies, one for each player. The expected payoff for player $i$ if all players play the strategy profile $\langle \sigma_i, \sigma_{-i} \rangle$ is $u_i(\sigma_i, \sigma_{-i})$, where $\sigma_{-i}$ denotes the strategies in $\sigma$ of all players other than $i$.
	
	Let $\pi^{\sigma}(h) = \prod_{h' \cdot a \sqsubseteq h} \sigma_{P(h')}(h',a)$ denote the probability of reaching $h$ if all players play according to $\sigma$. $\pi^{\sigma}_i(h)$ is the contribution of player $i$ to this probability (that is, the probability of reaching $h$ if chance and all players other than $i$ always chose actions leading to $h$). $\pi^{\sigma}_{-i}(h)$ is the contribution of all players, and chance, \emph{other than} $i$.
	We similarly define $\pi^{\sigma}(h,h')$ is the probability of reaching $h'$ given that $h$ has been reached, and $0$ if $h \not \sqsubset h'$. This papers focuses on \emph{perfect-recall} games, where a player never forgets past information. Thus, for every $I_i$, $\forall h, h' \in I_i$, $\pi^{\sigma}_i(h) = \pi^{\sigma}_i(h')$.
	We define $\pi^{\sigma}_i(I_i) = \pi^{\sigma}_i(h)$ for $h \in I_i$. Also, $I'_i \sqsubset I_i$ if for some $h' \in I'_i$ and some $h \in I_i$, $h' \sqsubset h$. Similarly, $I'_i \cdot a \sqsubset I_i$ if $h' \cdot a \sqsubset h$.
	
	A \emph{Nash equilibrium}~\cite{Nash50:Eq} is a strategy profile $\sigma^*$ where no player can improve by shifting to a different strategy, so $\sigma^*$ satisfies
	$\forall i,\ u_i(\sigma^*_i, \sigma^*_{-i}) = \max_{\sigma'_i \in \Sigma_i}u_i(\sigma'_i, \sigma^*_{-i})$.
	An \emph{$\epsilon$-Nash equilibrium} is a strategy profile $\sigma^*$ such that $\forall i,\ u_i(\sigma^*_i, \sigma^*_{-i}) + \epsilon \ge \max_{\sigma'_i \in \Sigma_i}u_i(\sigma'_i, \sigma^*_{-i})$.
	A \emph{best response} $BR(\sigma_{-i})$ is a strategy for player $i$ that is optimal against $\sigma_{-i}$. Formally, $BR(\sigma_{-i})$ satisfies $u_i(BR(\sigma_{-i}), \sigma_{-i}) = \max_{\sigma'_i \in \Sigma_i} u_i(\sigma_i', \sigma_{-i})$. In a two-player zero-sum game, the \emph{exploitability} $\textit{exp}(\sigma_{i})$ of a strategy $\sigma_{i}$ is how much worse $\sigma_i$ does against an opponent best response than a Nash equilibrium strategy would do. Formally, exploitability of $\sigma_i$ is $u_i(\sigma^*) - u_i(\sigma_i, BR(\sigma_{i}))$, where $\sigma^*$ is a Nash equilibrium.
	
	The expected \emph{value} of a node $h$ when players play according to $\sigma$ is $v_i^{\sigma}(h) = \sum_{z \in Z}\big(\pi^{\sigma}(h,z)u_i(z)\big)$. An infoset's value is the weighted average of the values of the nodes in the infoset, where a node is weighed by the player's belief that she is in that node. Formally, $v_i^{\sigma}(I_i) = \frac{\sum_{h \in I_i} \big(\pi_{-i}^{\sigma}(h) v_i^{\sigma}(h)\big)}{\sum_{h \in I_i} \pi_{-i}^{\sigma}(h)}$ and $v_i^{\sigma}(I_i,a) = \frac{\sum_{h \in I_i} \big(\pi_{-i}^{\sigma}(h) v_i^{\sigma}(h \cdot a)\big)}{\sum_{h \in I_i} \pi_{-i}^{\sigma}(h)}$.
	A \emph{counterfactual best response}~\cite{Moravcik16:Refining} $CBR(\sigma_{-i})$ is a best response that also maximizes value in unreached infosets. Specifically, a counterfactual best response is a best response $\sigma_i$ with the additional condition that if $\sigma_i(I_i,a) > 0$ then $v_i^{\sigma}(I_i,a) = \max_{a'} v_i^{\sigma}(I_i,a')$.
	We further define \emph{counterfactual best response value} $CBV^{\sigma_{-i}}(I_i)$ as the value player $i$ expects to achieve by playing according to $CBR(\sigma_{-i})$, having already reached infoset $I_i$.
	Formally, $CBV^{\sigma_{-i}}(I_i) = v_i^{\langle CBR({\sigma_{-i}}), \sigma_{-i} \rangle}(I_i)$ and $CBV^{\sigma_{-i}}(I_i,a) = v_i^{\langle CBR({\sigma_{-i}}), \sigma_{-i} \rangle}(I_i,a)$.
	
	An \emph{imperfect-information subgame}, which we refer to simply as a \emph{subgame} in this paper, can in most cases (but not all) be described as including all nodes which share prior \emph{public} actions (that is, actions viewable to both players). In poker, for example, a subgame is uniquely defined by a sequence of bets and public board cards. Figure~\ref{fig:cointoss}b shows the public game tree of Coin Toss.
	Formally, an imperfect-information subgame is a set of nodes $S \subseteq H$ such that for all $h \in S$, if $h \sqsubset h'$, then $h' \in S$, and for all $h \in S$ and all $i \in \mathcal{P}$, if $h' \in I_i(h)$ then $h' \in S$. Define $S_{\textit{top}}$ as the set of earliest-reachable nodes in $S$. That is, $h \in S_{\textit{top}}$ if $h \in S$ and $h' \not \in S$ for any $h' \sqsubset h$.
	
	\section{Prior Approaches to Subgame Solving}
	\label{sec:prior}
	
	This section reviews prior techniques for subgame solving in imperfect-information games, which we build upon. Throughout this section, we refer to the Coin Toss game shown in Figure~\ref{fig:cointoss}a.
	
	As discussed in Section~\ref{sec:intro}, a standard approach to dealing with large imperfect-information games is to solve an abstraction of the game. The abstract solution is a (probably suboptimal) strategy profile in the full game. We refer to this full-game strategy profile as the blueprint. The goal of subgame solving is to improve upon the blueprint by changing the strategy only in a subgame.
	While the blueprint is frequently a Nash equilibrium (or approximate Nash equilibrium) in some abstraction of the full game, our techniques do not assume this. The blueprint can in fact be any arbitrary strategy in the full game.
	
	\begin{figure}[!h]
		\centering
		\includegraphics[width=60mm]{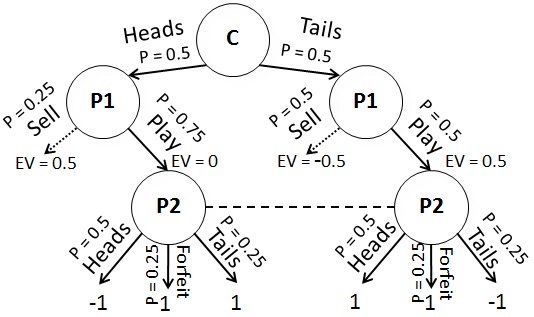}
		\caption{\small{The blueprint we refer to in the game of Coin Toss. The Sell action leads to a subgame that is not displayed. Probabilities are shown for all actions. The dotted line means the two $P_2$ nodes share an infoset. The EV of each $P_1$ action is also shown.}}
		\label{fig:blueprint}
	\end{figure}
	
	Assume that a blueprint $\sigma$ (shown in Figure~\ref{fig:blueprint}) has already been computed for Coin Toss in which $P_1$ chooses Play $\frac{3}{4}$ of the time with Heads and $\frac{1}{2}$ of the time with Tails, and $P_2$ chooses Heads $\frac{1}{2}$ of the time, Tails $\frac{1}{4}$ of the time, and Forfeit $\frac{1}{4}$ of the time after $P_1$ chooses Play.
	\footnote{In many large games the blueprint is far from optimal either because the equilibrium-finding algorithm did not sufficiently converge or because the game was too large and had to be abstracted. Clearly the example blueprint shown here could be trivially improved; we use it for simplicity of exposition.}
	The details of the blueprint in the Sell subgame are not relevant in this section, but the EV for choosing the Sell action \emph{is} relevant. We assume that if $P_1$ chose the Sell action and played optimally thereafter, then she would receive an expected payoff of $0.5$ if the coin is Heads, and $-0.5$ if the coin is Tails. We will attempt to improve $P_2$'s strategy in the subgame $S$ that follows $P_1$ choosing Play.
	
	\subsection{Unsafe Subgame Solving}
	We first review the most intuitive form of subgame solving, which we refer to as \emph{Unsafe subgame solving}~\cite{Billings03:Approximating,Gilpin06:Competitive,Gilpin07:Better,Ganzfried15:Endgame}.
	This form of subgame solving assumes both players played according to the blueprint prior to reaching the subgame. That defines a probability distribution over the nodes at the root of the subgame $S$, representing the probability that the true game state matches that node. A strategy for the subgame is then calculated which assumes that this distribution is correct.
	
\begin{figure}
		\centering
		\begin{subfigure}[t]{60mm}
			\centering
			\includegraphics[width=49mm]{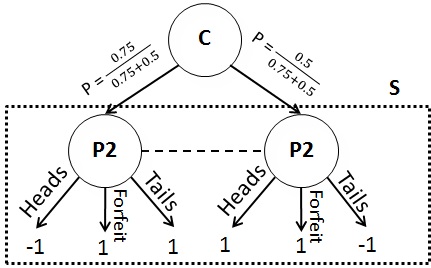}
			\caption{\small{Unsafe subgame solving}}\label{fig:unsafe}		
		\end{subfigure}
		\quad
		\begin{subfigure}[t]{72mm}
			\centering
			\includegraphics[width=62mm]{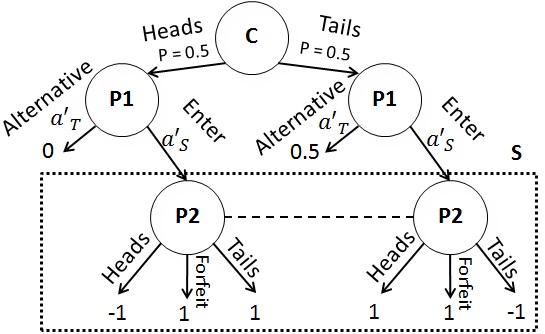}
			\caption{\small{Resolve subgame solving}}\label{fig:resolve}
		\end{subfigure}
		\caption{\small{The augmented subgames solved to find a $P_2$ strategy in the Play subgame of Coin Toss.}}\label{fig:1}
	\end{figure}

	In all subgame solving algorithms, an \emph{augmented subgame} containing $S$ and a few additional nodes is solved to determine the strategy for $S$. Applying Unsafe subgame solving to the blueprint in Coin Toss (after $P_1$ chooses Play) means solving the augmented subgame shown in Figure~\ref{fig:unsafe}.
	
	Specifically, the augmented subgame consists of only an initial chance node and $S$. The initial chance node reaches $h \in S_{\textit{top}}$ with probability $\frac{\pi^{\sigma}(h)}{\sum_{h' \in S_{\textit{top}}}\pi^{\sigma}(h')}$. The augmented subgame is solved and its strategy for $P_2$ is used in $S$ rather than the blueprint strategy.
	
	Unsafe subgame solving lacks theoretical solution quality guarantees and there are many situations where it performs extremely poorly.
	Indeed, if it were applied to the blueprint of Coin Toss then $P_2$ would always choose Heads---which $P_1$ could exploit severely by only choosing Play with Tails. Despite the lack of theoretical guarantees and potentially bad performance, Unsafe subgame solving is simple and can \emph{sometimes} produce low-exploitability strategies, as we show later.
	
	We now move to discussing \emph{safe} subgame-solving techniques, that is, ones that ensure that the exploitability of the strategy is no higher than that of the blueprint strategy.
	
	\subsection{Subgame Resolving}
	In \emph{subgame Resolving}~\cite{Burch14:Solving}, a safe strategy is computed for $P_2$ in the subgame by solving the augmented subgame shown in Figure~\ref{fig:resolve}, producing an equilibrium strategy $\sigma^S$. This augmented subgame differs from Unsafe subgame solving by giving $P_1$ the option to ``opt out'' from entering $S$ and instead receive the EV of playing optimally against $P_2$'s blueprint strategy in $S$.
	
	Specifically, the augmented subgame for Resolving differs from unsafe subgame solving as follows. For each $h_{\textit{top}} \in S_{\textit{top}}$ we insert a new $P_1$ node $h_r$, which exists only in the augmented subgame, between the initial chance node and $h_{\textit{top}}$. The set of these $h_r$ nodes is $S_r$. The initial chance node connects to each node $h_r \in S_r$ in proportion to the probability that player $P_1$ could reach $h_{\textit{top}}$ if $P_1$ tried to do so (that is, in proportion to $\pi_{-1}^{\sigma}(h_{\textit{top}})$). At each node $h_r \in S_r$, $P_1$ has two possible actions. Action $a'_S$ leads to $h_{\textit{top}}$, while action $a'_T$ leads to a terminal payoff that awards the value of playing optimally against $P_2$'s blueprint strategy, which is $CBV^{\sigma_2}(I_1(h_{\textit{top}}))$. In the blueprint of Coin Toss, $P_1$ choosing Play after the coin lands Heads results in an EV of $0$, and $\frac{1}{2}$ if the coin is Tails. Therefore, $a'_T$ leads to a terminal payoff of $0$ for Heads and $\frac{1}{2}$ for Tails. After the equilibrium strategy $\sigma^S$ is computed in the augmented subgame, $P_2$ plays according to the computed subgame strategy $\sigma^S_2$ rather than the blueprint strategy when in $S$. The $P_1$ strategy $\sigma^S_1$ is not used.
	
	Clearly $P_1$ cannot do worse than always picking action $a'_T$ (which awards the highest EV $P_1$ could achieve against $P_2$'s blueprint). But $P_1$ also cannot do \emph{better} than always picking $a'_T$, because $P_2$ could simply play according to the blueprint in $S$, which means action $a'_S$ would give the same EV to $P_1$ as action $a'_T$ (if $P_1$ played optimally in $S$). In this way, the strategy for $P_2$ in $S$ is pressured to be no worse than that of the blueprint.
	In Coin Toss, if $P_2$ were to always choose Heads (as was the case in Unsafe subgame solving), then $P_1$ would always choose $a'_T$ with Heads and $a'_S$ with Tails.
	
	Resolving guarantees that $P_2$'s exploitability will be no higher than the blueprint's (and may be better).
	However, it may miss opportunities for improvement. For example, if we apply Resolving to the example blueprint in Coin Toss, one solution to the augmented subgame is the blueprint itself, so $P_2$ may choose Forfeit $25\%$ of the time even though Heads and Tails dominate that action.
	Indeed, the original purpose of Resolving was not to \emph{improve} upon a blueprint strategy in a subgame, but rather to compactly store it by keeping only the EV at the root of the subgame and then reconstructing the strategy in real time when needed rather than storing the whole subgame strategy.
	
	Maxmargin subgame solving~\cite{Moravcik16:Refining}, discussed in Appendix~\ref{sec:maxmargin}, can improve performance by defining a \emph{margin}
	$M^{\sigma^S}(I_1) = CBV^{\sigma_2}(I_1) - CBV^{\sigma^S_2}(I_1)$
	for each $I_1 \in S_{\textit{top}}$ and maximizing $\min_{I_1 \in S_{\textit{top}}}M^{\sigma^S}(I_1)$. Resolving only makes all margins nonnegative. However, Maxmargin does worse in practice when using estimates of equilibrium values as discussed in Section~\ref{sec:error}.
	
	\section{Reach Subgame Solving}
	\label{sec:reachmaxmargin}
	
	All of the subgame-solving techniques described in Section~\ref{sec:prior} only consider the target subgame in isolation, which can lead to suboptimal strategies. For example, Maxmargin solving applied to $S$ in Coin Toss results in $P_2$ choosing Heads with probability $\frac{5}{8}$ and Tails with $\frac{3}{8}$ in $S$. This results in $P_1$ receiving an EV of $-\frac{1}{4}$ by choosing Play in the Heads state, and an EV of $\frac{1}{4}$ in the Tails state. However, $P_1$ could simply always choose Sell in the Heads state (earning an EV of $0.5$) and Play in the Tails state and receive an EV of $\frac{3}{8}$ for the entire game. In this section we introduce \emph{Reach subgame solving}, an improvement to past subgame-solving techniques that considers \emph{what the opponent could have alternatively received from other subgames}.\footnote{Other subgame-solving methods have also considered the cost of reaching a subgame~\cite{Waugh09:Strategy,Jackson14:Time}. However, those approaches are not correct in theory when applied in real time to any subgame reached during play.}
	For example, a better strategy for $P_2$ would be to choose Heads with probability $\frac{3}{4}$ and Tails with probability $\frac{1}{4}$. Then $P_1$ is indifferent between choosing Sell and Play in both cases and overall receives an expected payoff of $0$ for the whole game.
	
	However, that strategy is only optimal if $P_1$ would indeed achieve an EV of $0.5$ for choosing Sell in the Heads state and $-0.5$ in the Tails state. That would be the case if $P_2$ played according to the blueprint in the Sell subgame (which is not shown), but in reality we would apply subgame solving to the Sell subgame if the Sell action were taken, which would change $P_2$'s strategy there and therefore $P_1$'s EVs. Applying subgame solving to any subgame encountered during play is equivalent to applying it to all subgames independently. Thus, we must consider that the EVs from other subgames may differ from what the blueprint says because subgame solving would be applied to them as well.
	
	\begin{figure}[!h]
		\centering
		\includegraphics[width=130mm]{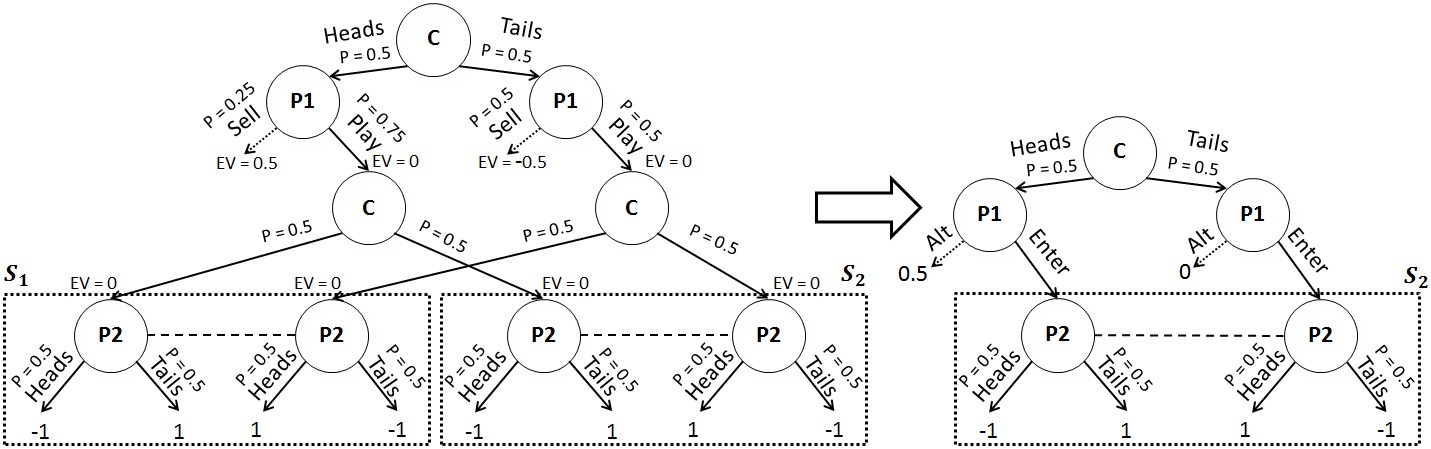}
		\caption{\small{Left: A modified game of Coin Toss with two subgames. The nodes $C_1$ and $C_2$ are public chance nodes whose outcomes are seen by both $P_1$ and $P_2$. Right: An augmented subgame for one of the subgames according to Reach subgame solving. If only one of the subgames is being solved, then the alternative payoff for Heads can be at most $1$. However, if both are solved independently, then the gift must be split among the subgames and must sum to at most $1$. For example, the alternative payoff in both subgames can be $0.5$.}}
		\label{fig:reach}
	\end{figure}
	
	As an example of this issue, consider the game shown in Figure~\ref{fig:reach} which contains two identical subgames $S_1$ and $S_2$ where the blueprint has $P_2$ pick Heads and Tails with 50\% probability. The Sell action leads to an EV of $0.5$ from the Heads state, while Play leads to an EV of $0$. If we were to solve just $S_1$, then $P_2$ could afford to always choose Tails in $S_1$, thereby letting $P_1$ achieve an EV of $1$ for reaching that subgame from Heads because, due to the chance node $C_1$, $S_1$ is only reached with 50\% probability. Thus, $P_1$'s EV for choosing Play would be $0.5$ from Heads and $-0.5$ from Tails, which is optimal. We can achieve this strategy in $S_1$ by solving an augmented subgame in which the alternative payoff for Heads is $1$. In that augmented subgame, $P_2$ always choosing Tails would be a solution (though not the only solution).
	
	However, if the same reasoning were applied independently to $S_2$ as well, then $P_2$ might always choose Tails in both subgames and $P_1$'s EV for choosing Play from Heads would become $1$ while the EV for Sell would only be $0.5$. Instead, we could allow $P_1$ to achieve an EV of $0.5$ for reaching each subgame from Heads (by setting the alternative payoff for Heads to $0.5$). In that case, $P_1$'s overall EV for choosing Play could only increase to $0.5$, even if both $S_1$ and $S_2$ were solved independently.
	
	We capture this intuition by considering for each $I_1 \in S_{\textit{top}}$ all the infosets and actions $I'_1 \cdot a' \sqsubset I_1$ that $P_1$ would have taken along the path to $I_1$. If, at some $I'_1 \cdot a' \sqsubset I_1$ where $P_1$ acted, there was a different action $a^* \in A(I'_1)$ that leads to a higher EV, then $P_1$ would have taken a suboptimal action if they reached $I_1$. The difference in value between $a^*$ and $a'$ is referred to as a \emph{gift}. We can afford to let $P_1$'s value for $I_1$ increase beyond the blueprint value (and in the process lower $P_1$'s value in some other infoset in $S_{\textit{top}}$), so long as the increase to $I_1$'s value is small enough that choosing actions leading to $I_1$ is still suboptimal for $P_1$. Critically, we must ensure that the increase in value is small enough even when the potential increase across all subgames is summed together, as in Figure~\ref{fig:reach}.\footnote{In this paper and in our experiments, we allow any infoset that descends from a gift to increase by the size of the gift (e.g., in Figure~\ref{fig:reach} the gift from Heads is $0.5$, so we allow $P_1$'s value for Heads in both $S_1$ and $S_2$ to increase by $0.5$). However, any division of the gift among subgames is acceptable so long as the potential increase across all subgames (multiplied by the probability of $P_1$ reaching that subgame) does not exceed the original gift. For example in Figure~\ref{fig:reach} if we only apply Reach subgame solving to $S_1$, then we could allow the Heads state in $S_1$ to increase by $1$ rather than just by $0.5$. In practice, some divisions may do better than others. The division we use in this paper (applying gifts equally to all subgames) did well in practice.}
	
	A complicating factor is that gifts we assumed were present may actually not exist. For example, in Coin Toss, suppose applying subgame solving to the Sell subgame results in $P_1$'s value for Sell from the Heads state decreasing from $0.5$ to $0.25$. If we independently solve the Play subgame, we have no way of knowing that $P_1$'s value for Sell is lower than the blueprint suggested, so we may still assume there is a gift of $0.5$ from the Heads state based on the blueprint. Thus, in order to guarantee a theoretical result on exploitability that is as strong as possible, we use in our theory and experiments a \emph{lower bound} on what gifts could be after subgame solving was applied to all other subgames.
	
	Formally, let $\sigma_2$ be a $P_2$ blueprint and let $\sigma^{-S}_2$ be the $P_2$ strategy that results from applying subgame solving independently to a set of disjoint subgames other than $S$. Since we do not want to compute $\sigma_2^{-S}$ in order to apply subgame solving to $S$, let $\lfloor g^{\sigma_2^{-S}}(I'_1,a') \rfloor$ be a lower bound of $CBV^{\sigma^{-S}_2}(I'_1) - CBV^{\sigma^{-S}_2}(I'_1,a')$ that does not require knowledge of $\sigma_2^{-S}$. In our experiments we use $\lfloor g^{\sigma_2^{-S}}(I'_1,a') \rfloor = \max_{a \in A_z(I'_1) \cup \{a'\}} CBV^{\sigma_2}(I'_1,a) - CBV^{\sigma_2}(I'_1,a')$ where $A_z(I'_1) \subseteq A(I'_1)$ is the set of actions leading immediately to terminal nodes.
	Reach subgame solving modifies the augmented subgame in Resolving and Maxmargin by increasing the alternative payoff for infoset $I_1 \in S_{\textit{top}}$ by $\sum_{I'_1 \cdot a' \sqsubseteq I_1 \mid P(I'_1) = P_1} \lfloor g^{\sigma_2^{-S}}(I'_1,a') \rfloor$.
	Formally, we define a \emph{reach margin} as 
	\begin{equation}
	M_r^{\sigma^S}(I_1) = M^{\sigma^S}(I_1) + \sum_{I'_1 \cdot a' \sqsubseteq I_1 \mid P(I'_1) = P_1} \lfloor g^{\sigma_2^{-S}}(I'_1,a') \rfloor
	\end{equation}
	This margin is larger than or equal to the one for Maxmargin, because $\lfloor g^{\sigma_2^{-S}}(I',a') \rfloor$ is nonnegative. We refer to the improved algorithms as Reach-Resolve and Reach-Maxmargin.
	
	Intuitively, the alternative payoff in an augmented subgame determines how important it is that $P_2$ ``defend'' against that $P_1$ infoset. If the alternative payoff is increased, then $P_1$ is more likely to choose the alternative payoff rather than enter the subgame, so $P_2$ can instead focus on lowering the value of other $P_1$ infosets in $S_{\textit{top}}$.
	
	Using a lower bound on gifts is not necessary to guarantee safety. So long as we use a gift value $g^{\sigma'}(I'_1,a') \le CBV^{\sigma_2}(I'_1) - CBV^{\sigma_2}(I'_1,a')$, the resulting strategy will be safe. However, using a lower bound further guarantees a reduction to exploitability when a $P_1$ best response reaches with positive probability an infoset $I_1 \in S_{\textit{top}}$ that has positive margin, as proven in Theorem~\ref{th:multiple}. In practice, it may be best to use an accurate estimate of gifts. One option is to use 
	$
	\hat{g}^{\sigma_2^{-S}}(I'_1,a') = \tilde{CBV}^{\sigma_2}(I'_1) - \tilde{CBV}^{\sigma_2}(I'_1,a')
	$
	in place of $\lfloor g^{\sigma_2^{-S}}(I'_1,a') \rfloor$, where $\tilde{CBV}^{\sigma_2}$ is the closest $P_1$ can get to the value of a counterfactual best response while $P_1$ is constrained to playing within the abstraction that generated the blueprint. Using estimates is covered in more detail in Section~\ref{sec:error}.
	
	Theorem~\ref{th:multiple} shows that when subgames are solved independently and using lower bounds on gifts, Reach-Maxmargin solving has exploitability lower than or equal to past safe techniques. The theorem statement is similar to that of Maxmargin~\cite{Moravcik16:Refining}, but the margins are now larger (or equal) in size.
	\begin{theorem}
		Given a strategy $\sigma_2$ in a two-player zero-sum game, a set of disjoint subgames $\mathbb{S}$, and a strategy $\sigma_2^S$ for each subgame $S \in \mathbb{S}$ produced via Reach-Maxmargin solving using lower bounds for gifts, let $\sigma'_2$ be the strategy that plays according to $\sigma^S_2$ for each subgame $S \in \mathbb{S}$, and  $\sigma_2$ elsewhere. Moreover, let $\sigma_2^{-S}$ be the strategy that plays according to $\sigma'_2$ everywhere except for $P_2$ nodes in $S$, where it instead plays according to $\sigma_2$. If $\pi_1^{BR(\sigma'_2)}(I_1) > 0$ for some $I_1 \in S_{\textit{top}}$, then $\textit{exp}(\sigma'_2) \le \textit{exp}(\sigma^{-S}_2) - \sum_{h \in I_1} \pi_{-1}^{\sigma_2}(h) M_{r}^{\sigma^S}(I_1)$.
		\label{th:multiple}
	\end{theorem}

	So far the described techniques have guaranteed a reduction in exploitability over the blueprint by setting the value of $a'_T$ equal to the value of $P_1$ playing optimally to $P_2$'s blueprint. Relaxing this guarantee by instead setting the value of $a'_T$ equal to an \emph{estimate} of $P_1$'s value when \emph{both} players play optimally leads to far lower exploitability in practice. We discuss this approach in the next section.
	
	\section{Estimates for Alternative Payoffs}
	\label{sec:error}
	
	In this section we consider the case where we have a good estimate of what the values of subgames would look like in a Nash equilibrium.
	Unlike previous sections, exploitability might be \emph{higher} than the blueprint when using this method; the solution quality ultimately depends on the accuracy of the estimates used. In practice this approach leads to significantly lower exploitability.
	
	When solving multiple $P_2$ subgames, there is a minimally-exploitable strategy $\sigma^*_2$ that could, in theory, be computed by changing only the strategies in the subgames. ($\sigma^*_2$ may not be a Nash equilibrium because $P_2$'s strategy outside the subgames is fixed, but it is the closest that can be achieved by changing the strategy only in the subgames). However, $\sigma^*_2$ can only be guaranteed to be produced by solving all the subgames together, because the optimal strategy in one subgame depends on the optimal strategy in other subgames.
	
	Still, suppose that we know $CBV^{\sigma^*_2}(I_1)$ for every infoset $I_1 \in S_{\textit{top}}$ for every subgame $S$. Let $I_{r,1}$ be the infoset in $S_r$ that leads to $I_1$. By setting the $P_1$ alternative payoff for $I_{r,1}$ to $v(I_{r,1},a'_T) = CBV^{\sigma^*_2}(I_1)$, safe subgame solving guarantees a strategy will be produced with exploitability no worse than $\sigma^*_2$. Thus, achieving a strategy equivalent to $\sigma^*_2$ does not require knowledge of $\sigma^*_2$; rather, it only requires knowledge of $CBV^{\sigma^*_2}(I_1)$ for infosets $I_1$ in the top of the subgames.
	
	While we do not know $CBV^{\sigma^*_2}(I_1)$ exactly without knowing $\sigma^*_2$ itself, we may nevertheless be able to produce (or learn) good \emph{estimates} of $CBV^{\sigma^*_2}(I_1)$. For example, in Section~\ref{sec:results} we compute the solution to the game of No-Limit Flop Hold'em (NLFH), and find that in perfect play $P_2$ can expect to win about $37$ mbb/h\footnote{In poker, the performance of one strategy against another depends on how much money is being wagered. For this reason, expected value and exploitability are measured in milli big blinds per hand (mbb/h). A big blind is the amount of money one of the players is required to put into the pot at the beginning of each hand.} (that is, if $P_1$ plays perfectly against the computed $P_2$ strategy, then $P_1$ earns $-37$; if $P_2$ plays perfectly against the computed $P_1$ strategy, then $P_2$ earns $37$). An abstraction of the game which is only $0.02\%$ of the size of the full game produces a $P_1$ strategy that can be beaten by $112$ mbb/h, and a $P_2$ strategy that can be beaten by $21$ mbb/h. Still, the abstract strategy estimates that at equilibrium, $P_2$ can expect to win $35$ mbb/h. So even though the abstraction produces a very poor estimate of the \emph{strategy} $\sigma^*$, it produces a good estimate of the \emph{value} of $\sigma^*$. In our experiments, we estimate $CBV^{\sigma^*_2}(I_1)$ by calculating a $P_1$ counterfactual best response \emph{within the abstract game} to $P_2$'s blueprint. We refer to this strategy as $\tilde{CBR}(\sigma_2)$ and its value in an infoset $I_1$ as $\tilde{CBV}^{\sigma_2}(I_1)$. We then use $\tilde{CBV}^{\sigma_2}(I_1)$ as the alternative payoff of $I_1$ in an augmented subgame. In other words, rather than calculate a $P_1$ counterfactual best response in the full game to $P_2$'s blueprint strategy (which would be $CBR(\sigma_2)$), we instead calculate $P_1$'s counterfactual best response where $P_1$ is constrained by the abstraction.
	
	If the blueprint was produced by conducting $T$ iterations of CFR in an abstract game, then one could instead simply use the final iteration's strategy $\sigma_1^T$, as this converges to a counterfactual best response within the abstract game. This is what we use in our experiments in this paper.
	
	Theorem~\ref{th:estimate} proves that if we use estimates of $CBV^{\sigma^*_2}(I_1)$ as the alternative payoffs in Maxmargin subgame solving, then we can bound exploitability by the distance of the estimates from the true values. This is in contrast to the previous algorithms which guaranteed exploitability no worse than the blueprint.
	
	\begin{theorem}
		\label{th:estimate}
		Let $\mathbb{S}$ be a set of disjoint subgames being solved in a game with no private actions. Let $\sigma$ be a blueprint and let $\sigma_2^*$ be a minimally-exploitable $P_2$ strategy that differs from $\sigma_2$ only in $\mathbb{S}$. Let $\Delta = \max_{S \in \mathbb{S}, I_1 \in S_{\textit{top}}} |CBV^{\sigma^*_2}(I_1) - CBV^{\sigma_2}(I_1)|$. Applying Maxmargin solving to each subgame using $\sigma$ as the blueprint produces a $P_2$ strategy with exploitability no higher than $exp(\sigma^*_2) + 2\Delta$.
	\end{theorem}

	Using estimates of the values of $\sigma^*$ tends to be do better than the theoretically safe options described in Section~\ref{sec:prior}.\footnote{It is also possible to combine the safety of past approaches with some of the better performance of using estimates by adding the original Resolve conditions as additional constraints.}
	
	Although Theorem~\ref{th:estimate} uses Maxmargin in the proof, in practice Resolve does far better with estimates than Maxmargin. Additionally, the theorem easily extends to Reach-Maxmargin as well, and Reach-Resolve does better than Resolve regardless of whether estimates are used.
	
	Section~\ref{sec:distribution} discusses an improvement, which we refer to as Distributional alternative payoffs, that leads to even better performance by making the algorithm more robust to errors in the blueprint estimates.
	
	\section{Nested Subgame Solving}
	\label{sec:iterative}
	
	As we have discussed, large games must be abstracted to reduce the game to a tractable size. This is particularly common in games with large or continuous action spaces.
	Typically the action space is discretized by action abstraction so that only a few actions are included in the abstraction. While we might limit ourselves to the actions we included in the abstraction, an opponent might choose actions that are not in the abstraction. In that case, the \emph{off-tree} action can be mapped to an action that is in the abstraction, and the strategy from that in-abstraction action can be used.
	For example, in an auction game we might include a bid of $\$100$ in our abstraction. If a player bids $\$101$, we simply treat that as a bid of $\$100$. This is referred to as \emph{action translation}~\cite{Gilpin08:Heads-up,Schnizlein09:Probabilistic,Ganzfried13:Action}. Action translation is the state-of-the-art prior approach to dealing with this issue.  It has been used, for example, by all the leading competitors in the Annual Computer Poker Competition (ACPC).
	
	In this section, we develop techniques for applying subgame solving to calculate responses to opponent off-tree actions, thereby obviating the need for action translation. That is, rather than simply treat a bid of $\$101$ as $\$100$, we calculate in real time a unique response to the bid of $\$101$. This can also be done in a nested fashion in response to subsequent opponent off-tree actions.
	We present two methods that dramatically outperform the leading action translation technique.
	Additionally, these techniques can be used to solve finer-grained models as play progresses down the game tree. For exposition, we assume that $P_2$ wishes to respond to $P_1$ choosing an off-tree action.
	
	We refer to the first method as the \emph{inexpensive} method.\footnote{Following our study, the AI DeepStack used a technique similar to this form of nested subgame solving~\cite{Moravcik17:DeepStack}.} When $P_1$ chooses an off-tree action $a$, a subgame $S$ is generated following that action such that for any infoset $I_1$ that $P_1$ might be in, $I_1 \cdot a \in S_{\textit{top}}$. This subgame may itself be an abstraction. A solution $\sigma^S$ is computed via subgame solving, and $\sigma^S$ is combined with $\sigma$ to form a new blueprint $\sigma'$ in the expanded abstraction that now includes action $a$. The process repeats whenever $P_1$ again chooses an off-tree action.
	
	To conduct safe subgame solving in response to off-tree action~$a$, we could calculate $CBV^{\sigma_2}(I_1,a)$ by defining, via action translation, a $P_2$ blueprint following $a$ and best responding to it~\cite{Brown15:Simultaneous}. However, that could be computationally expensive and would likely perform poorly in practice because, as we show later, action translation is highly exploitable. Instead, we relax the guarantee of safety and use $\tilde{CBV}^{\sigma_2}(I_1)$ for the alternative payoff, where $\tilde{CBV}^{\sigma_2}(I_1)$ is the value in $I_1$ of $P_1$ playing as close to optimal as possible while constrained to playing in the blueprint abstraction (which excludes action~$a$).
	In this case, exploitability depends on how well $\tilde{CBV}^{\sigma_2}(I_1)$ approximates $CBV^{\sigma^*_2}(I_1)$, where $\sigma^*_2$ is an optimal $P_2$ strategy (see Section~\ref{sec:error}).\footnote{We estimate $CBV^{\sigma^*_2}(I_1)$ rather than $CBV^{\sigma^*_2}(I_1,a)$ because $CBV^{\sigma^*_2}(I_1) - CBV^{\sigma^*_2}(I_1,a)$ is a gift that may be added to the alternative payoff anyway.}
	In general, we find that only a small number of near-optimal actions need to be included in the blueprint abstraction for $\tilde{CBV}^{\sigma_2}(I_1)$ to be close to $CBV^{\sigma^*_2}(I_1)$. We can then approximate a near-optimal response to any opponent action. This is particularly useful in very large or continuous action spaces.
	
	The ``inexpensive'' approach cannot be combined with Unsafe subgame solving because the probability of reaching an action outside of a player's abstraction is undefined.
	Nevertheless, a similar approach is possible with Unsafe subgame solving (as well as all the other subgame-solving techniques) by starting the subgame solving at $h$ rather than at $h \cdot a$. In other words, if action $a$ taken in node $h$ is not in the abstraction, then Unsafe subgame solving is conducted in the smallest subgame containing $h$ (and action $a$ is added to that abstraction). This increases the size of the subgame compared to the inexpensive method because a strategy must be recomputed for every action $a' \in A(h)$ in addition to $a$.
	For example, if an off-tree action is chosen by the opponent as the first action in the game, then the strategy for the entire game must be recomputed.
	We therefore call this method the \emph{expensive} method. We present experiments with both methods.
	
	\section{Experiments}
	\label{sec:results}
	
	Our experiments were conducted on heads-up no-limit Texas hold'em, as well as two smaller-scale poker games we call \emph{No-Limit Flop Hold'em} (NLFH) and \emph{No-Limit Turn Hold'em} (NLTH). The description for these games can be found in Appendix~\ref{sec:rules}. For equilibrium finding, we used CFR+~\cite{Tammelin15:Solving}.
	
	Our first experiment compares the performance of the subgame-solving techniques when applied to information abstraction (which is card abstraction in the case of poker). Specifically, we solve NLFH with no information abstraction on the preflop. On the flop, there are 1,286,792 infosets for each betting sequence; the abstraction buckets them into 200, 2,000, or 30,000 abstract ones (using a leading information abstraction algorithm~\cite{Ganzfried14:Potential-Aware}).  We then apply subgame solving immediately after the flop community cards are dealt.
	
	We experiment with two versions of the game, one small and one large, which include only a few of the available actions in each infoset. We also experimented on abstractions of NLTH. In that case, we solve NLTH with no information abstraction on the preflop or flop. On the turn, there are 55,190,538 infosets for each betting sequence; the abstraction buckets them into 200, 2,000, or 20,000 abstract ones. We apply subgame solving immediately after the turn community card is dealt.
	
	Tables~\ref{ta:smallflop}, \ref{ta:largeflop}, and \ref{ta:turn} show the performance of each technique. In all our experiments, exploitability is measured in the standard units used in this field: milli big blinds per hand (mbb/h).
	
	\begin{table}[!h]
		\centering
		\begin{tabular}{|l|l|l|l|}
			\hline
			\textbf{Small Flop Hold'em Flop Buckets:} & \textbf{200}  & \textbf{2,000} & \textbf{30,000} \\ \hline
			Trunk Strategy   & 88.69 & 37.374 & 9.128 \\ \hline
			Unsafe          & 14.68 & 3.958 & 0.5514 \\ \hline
			Resolve         & 60.16 & 17.79 & 5.407 \\ \hline
			Maxmargin       & 30.05 & 13.99 & 4.343 \\ \hline
			Reach-Maxmargin & 29.88 & 13.90 & 4.147 \\ \hline
			Reach-Maxmargin (not split) & 24.87 & 9.807 & 2.588 \\ \hline
			Estimate		& 11.66 & 6.261 & 2.423 \\ \hline
			Estimate + Distributional & 10.44 & 6.245 & 3.430 \\ \hline
			Reach-Estimate + Distributional & 10.21 & 5.798 & 2.258 \\ \hline
			Reach-Estimate + Distributional (not split) & 9.560 & 4.924 & 1.733 \\ \hline
		\end{tabular}
		\caption{Exploitability (evaluated in the game with no information abstraction) of subgame-solving in small flop Texas hold'em.}
		\label{ta:smallflop}
		\vspace{-0.1in}
	\end{table}
	
	\begin{table}[!h]
		\centering
		\begin{tabular}{|l|l|l|l|}
			\hline
			\textbf{Large Flop Hold'em Flop Buckets:} & \textbf{200}  & \textbf{2,000} & \textbf{30,000} \\ \hline
			Trunk Strategy   & 283.7 & 165.2 & 41.41 \\ \hline
			Unsafe          & 65.59 & 38.22 & 396.8 \\ \hline
			Resolve         & 179.6 & 101.7 & 23.11 \\ \hline
			Maxmargin       & 134.7 & 77.89 & 19.50 \\ \hline
			Reach-Maxmargin & 134.0 & 72.22 & 18.80 \\ \hline
			Reach-Maxmargin (not split) & 130.3 & 66.79 & 16.41 \\ \hline
			Estimate 		& 52.62 & 41.93 & 30.09 \\ \hline
			Estimate + Distributional & 49.56 & 38.98 & 10.54 \\ \hline
			Reach-Estimate + Distributional & 49.33 & 38.52 & 9.840 \\ \hline
			Reach-Estimate + Distributional (not split) & 49.13 & 37.22 & 8.777 \\ \hline
		\end{tabular}
		\caption{Exploitability (evaluated in the game with no information abstraction) of subgame-solving in large flop Texas hold'em.}
		\label{ta:largeflop}
		\vspace{-0.1in}
	\end{table}
	
	\begin{table}[!h]
		\centering
		\begin{tabular}{|l|l|l|l|}
			\hline
			\textbf{Turn Hold'em Turn Buckets:} & \textbf{200} & \textbf{2,000} & \textbf{20,000} \\ \hline
			Trunk Strategy   & 684.6 & 465.1 & 345.5 \\ \hline
			Unsafe          & 130.4 & 85.95 & 79.34 \\ \hline
			Resolve         & 454.9 & 321.5 & 251.8 \\ \hline
			Maxmargin       & 427.6 & 299.6 & 234.4 \\ \hline
			Reach-Maxmargin & 424.4 & 298.3 & 233.5 \\ \hline
			Reach-Maxmargin (not split) & 333.4 & 229.4 & 175.5 \\ \hline
			Estimate 		& 120.6 & 89.43 & 76.44 \\ \hline
			Estimate + Distributional & 119.4 & 87.83 & 74.35 \\ \hline
			Reach-Estimate + Distributional & 116.8 & 85.80 & 72.59 \\ \hline
			Reach-Estimate + Distributional (not split) & 113.3 & 83.24 & 70.68 \\ \hline
		\end{tabular}
		\caption{Exploitability (evaluated in the game with no information abstraction) of subgame-solving in turn Texas hold'em.}
		\label{ta:turn}
		\vspace{-0.1in}
	\end{table}
	
	In the above experiments, Estimate is the technique introduced in Section~\ref{sec:error} (added on top of Resolving) and Distributional is the technique introduced in Appendix~\ref{sec:distribution}. We use a normal distribution in the Distributional subgame solving experiments, with standard deviation determined by the heuristic presented in Appendix~\ref{sec:distribution}.
	
	Since subgame solving begins immediately after a chance node with an extremely high branching factor ($1,755$ in NLFH), the gifts for the Reach algorithms are divided among subgames inefficiently. Many subgames do not use the gifts at all, while others could make use of more.
	The result is that the theoretically safe version of Reach allocates gifts very conservatively.
	In the experiments we show results both for the theoretically safe splitting of gifts, as well as a more aggressive version where gifts are scaled up by the branching factor of the chance node ($1,755$). This weakens the theoretical guarantees of the algorithm, but in general did better than splitting gifts in a theoretically correct manner. However, this is not universally true. Appendix~\ref{sec:reachscale} shows that in at least one case, exploitability increased when gifts were scaled up too aggressively. In all cases, using Reach subgame solving in at least the theoretical safe method led to lower exploitability.
	
	Despite lacking theoretical guarantees, Unsafe subgame solving did surprisingly well in most games. However, it did substantially worse in Large NLFH with 30,000 buckets. This exemplifies its variability. Among the safe methods, all of the changes we introduce show improvement over past techniques. The Reach-Estimate + Distributional algorithm generally resulted in the lowest exploitability among the various choices, and in most cases beat Unsafe subgame solving.
	
	In all but one case, using estimated values lowered exploitability more than Maxmargin and Resolve subgame solving. Also, in all but one case using distributional alternative payoffs lowered exploitability.
	
	The second experiment evaluates nested subgame solving, and compares it to action translation. In order to also evaluate action translation, in this experiment, we create an NLFH game that includes 3 bet sizes at every point in the game tree (0.5, 0.75, and 1.0 times the size of the pot); a player can also decide not to bet. Only one bet (i.e., no raises) is allowed on the preflop, and three bets are allowed on the flop. There is no information abstraction anywhere in the game.
	We also created a second, smaller abstraction of the game in which there is still no information abstraction, but the 0.75$\times$ pot bet is never available. We calculate the exploitability of one player using the smaller abstraction, while the other player uses the larger abstraction. Whenever the large-abstraction player chooses a 0.75$\times$ pot bet, the small-abstraction player generates and solves a subgame for the remainder of the game (which again does not include any subsequent 0.75$\times$ pot bets) using the nested subgame-solving techniques described above. This subgame strategy is then used as long as the large-abstraction player plays within the small abstraction, but if she chooses the 0.75$\times$ pot bet again later, then the subgame solving is used again, and so on.
	
	Table~\ref{ta:iterative} shows that all the subgame-solving techniques substantially outperform action translation.
	Resolve, Maxmargin, and Reach-Maxmargin use inexpensive nested subgame solving, while Unsafe and ``Reach-Maxmargin (expensive)'' use the expensive approach. In all cases, we used estimates for the alternative payoff as described in Section~\ref{sec:iterative}.
	We did not test distributional alternative payoffs in this experiment, since the calculated best response values are likely quite accurate.
	Reach-Maxmargin performed the best, outperforming Maxmargin and Unsafe subgame solving.
	These results suggest that nested subgame solving is preferable to action translation (if there is sufficient time to solve the subgame).
	
	\begin{table}[!h]
		\centering
		\small{
			\begin{tabular}{|l|l|}
				\hline
				& \textbf{mbb/h} \\ \hline
				Randomized Pseudo-Harmonic Mapping          & 1,465 \\ \hline
				Resolve         & 150.2 \\ \hline
				Reach-Maxmargin (Expensive) & 149.2 \\ \hline
				Unsafe (Expensive) & 148.3 \\ \hline
				Maxmargin        & 122.0 \\ \hline
				Reach-Maxmargin  & 119.1 \\ \hline
			\end{tabular}
			\caption{\small{Exploitability of the various subgame-solving techniques in nested subgame solving. The performance of the pseudo-harmonic action translation is also shown.}}
			\label{ta:iterative}
		}
	\end{table}
	
	We used the techniques presented in this paper in our AI Libratus, which competed against four top human specialists in heads-up no-limit Texas hold'em in the January 2017 Brains vs. AI competition. Libratus was constructed by first solving an abstraction of the game via a new variant of Monte Carlo CFR~\cite{Lanctot09:Monte} that prunes negative-regret actions~\cite{Brown15:Regret-Based,Brown16:Baby,Brown16:Reduced}. Libratus applied nested subgame solving (solved with CFR+~\cite{Tammelin15:Solving}) upon reaching the third betting round, and in response to every subsequent opponent bet thereafter. This allowed Libratus to avoid information abstraction during play, and leverage nested subgame solving's far lower exploitability in response to opponent off-tree actions.
	
\begin{figure}[!h]
		\centering
		\includegraphics[width=67mm]{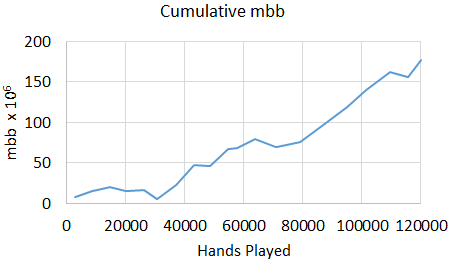}
		\hspace{4mm}
		\includegraphics[width=67mm]{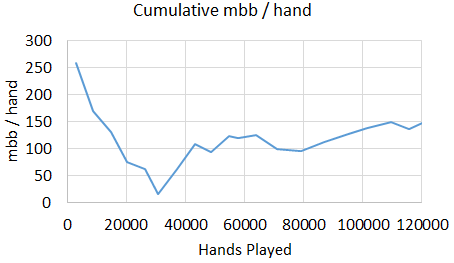}
		\caption{\emph{Libratus}'s performance over the course of the 2017 \emph{Brains vs AI} competition.}
		\label{fig:mbb}
	\end{figure}

	No-limit Texas hold'em is the most popular form of poker in the world and has been the primary benchmark challenge for AI in imperfect-information games. The competition was played over the course of 20 days, and involved 120,000 hands of poker. A prize pool of \$200,000 was split among the four humans based on their performance against the AI to incentivize strong play. The AI decisively defeated the team of human players by a margin of $147$ mbb / hand, with 99.98 statistical significance (see Figure~\ref{fig:mbb}).
	This was the first, and so far only, time an AI defeated top humans in no-limit poker.
	
	\section{Conclusion}
	
	We introduced a subgame-solving technique for imperfect-information games that has stronger theoretical guarantees and better practical performance than prior subgame-solving methods. We presented results on exploitability of both safe and unsafe subgame-solving techniques.  We also introduced a method for nested subgame solving in response to the opponent's off-tree actions, and demonstrated that this leads to dramatically better performance than the usual  approach of action translation. This is, to our knowledge, the first time that exploitability of subgame-solving techniques has been measured in large games.
	
	Finally, we demonstrated the effectiveness of these techniques in practice in heads-up no-limit Texas Hold'em poker, the main benchmark challenge for AI in imperfect-information games. We developed the first AI to reach the milestone of defeating top humans in heads-up no-limit Texas Hold'em.
	
	\newpage
	
	\section{Acknowledgments} This material is based on work supported by the National Science Foundation under grants IIS-1718457, IIS-1617590, and CCF-1733556, and the ARO under award W911NF-17-1-0082, as well as XSEDE computing resources provided by the Pittsburgh Supercomputing Center. The \emph{Brains vs. AI} competition was sponsored by Carnegie Mellon University, Rivers Casino, GreatPoint Ventures, Avenue4Analytics, TNG Technology Consulting, Artificial Intelligence, Intel, and Optimized Markets, Inc. We thank Kristen Gardner, Marcelo Gutierrez, Theo Gutman-Solo, Eric Jackson, Christian Kroer, Tim Reiff, and the anonymous reviewers for helpful feedback.
	
	\bibliographystyle{plain}

\begin{thebibliography}{10}
		
		\bibitem{Billings03:Approximating}
		Darse Billings, Neil Burch, Aaron Davidson, Robert Holte, Jonathan Schaeffer,
		Terence Schauenberg, and Duane Szafron.
		\newblock Approximating game-theoretic optimal strategies for full-scale poker.
		\newblock In {\em Proceedings of the 18th International Joint Conference on
			Artificial Intelligence (IJCAI)}, 2003.
		
		\bibitem{Bowling15:Heads-up}
		Michael Bowling, Neil Burch, Michael Johanson, and Oskari Tammelin.
		\newblock Heads-up limit hold'em poker is solved.
		\newblock {\em Science}, 347(6218):145--149, January 2015.
		
		\bibitem{Brown17:Dynamic}
		Noam Brown, Christian Kroer, and Tuomas Sandholm.
		\newblock Dynamic thresholding and pruning for regret minimization.
		\newblock In {\em AAAI Conference on Artificial Intelligence (AAAI)}, pages
		421--429, 2017.
		
		\bibitem{Brown15:Regret-Based}
		Noam Brown and Tuomas Sandholm.
		\newblock Regret-based pruning in extensive-form games.
		\newblock In {\em Advances in Neural Information Processing Systems}, pages
		1972--1980, 2015.
		
		\bibitem{Brown15:Simultaneous}
		Noam Brown and Tuomas Sandholm.
		\newblock Simultaneous abstraction and equilibrium finding in games.
		\newblock In {\em Proceedings of the International Joint Conference on
			Artificial Intelligence (IJCAI)}, 2015.
		
		\bibitem{Brown16:Baby}
		Noam Brown and Tuomas Sandholm.
		\newblock Baby {T}artanian8: Winning agent from the 2016 annual computer poker
		competition.
		\newblock In {\em Proceedings of the Twenty-Fifth International Joint
			Conference on Artificial Intelligence (IJCAI-16)}, pages 4238--4239, 2016.
		
		\bibitem{Brown16:Reduced}
		Noam Brown and Tuomas Sandholm.
		\newblock Reduced space and faster convergence in imperfect-information games
		via regret-based pruning.
		\newblock {\em arXiv preprint arXiv:1609.03234}, 2016.
		
		\bibitem{Burch14:Solving}
		Neil Burch, Michael Johanson, and Michael Bowling.
		\newblock Solving imperfect information games using decomposition.
		\newblock In {\em AAAI Conference on Artificial Intelligence (AAAI)}, pages
		602--608, 2014.
		
		\bibitem{Campbell02:Deep}
		Murray Campbell, A~Joseph Hoane, and Feng-{H}siung Hsu.
		\newblock Deep {B}lue.
		\newblock {\em Artificial intelligence}, 134(1-2):57--83, 2002.
		
		\bibitem{Ganzfried13:Action}
		Sam Ganzfried and Tuomas Sandholm.
		\newblock Action translation in extensive-form games with large action spaces:
		axioms, paradoxes, and the pseudo-harmonic mapping.
		\newblock In {\em Proceedings of the Twenty-Third international joint
			conference on Artificial Intelligence}, pages 120--128. AAAI Press, 2013.
		
		\bibitem{Ganzfried14:Potential-Aware}
		Sam Ganzfried and Tuomas Sandholm.
		\newblock Potential-aware imperfect-recall abstraction with earth mover's
		distance in imperfect-information games.
		\newblock In {\em AAAI Conference on Artificial Intelligence (AAAI)}, 2014.
		
		\bibitem{Ganzfried15:Endgame}
		Sam Ganzfried and Tuomas Sandholm.
		\newblock Endgame solving in large imperfect-information games.
		\newblock In {\em International Conference on Autonomous Agents and Multi-Agent
			Systems (AAMAS)}, pages 37--45, 2015.
		
		\bibitem{Gilpin12:First}
		Andrew Gilpin, Javier Pe{\~n}a, and Tuomas Sandholm.
		\newblock First-order algorithm with $\mathcal{O}(\mathrm{ln} (1/\epsilon))$
		convergence for $\epsilon$-equilibrium in two-person zero-sum games.
		\newblock {\em Mathematical Programming}, 133(1--2):279--298, 2012.
		\newblock Conference version appeared in AAAI-08.
		
		\bibitem{Gilpin06:Competitive}
		Andrew Gilpin and Tuomas Sandholm.
		\newblock A competitive {T}exas {H}old'em poker player via automated
		abstraction and real-time equilibrium computation.
		\newblock In {\em Proceedings of the National Conference on Artificial
			Intelligence (AAAI)}, pages 1007--1013, 2006.
		
		\bibitem{Gilpin07:Better}
		Andrew Gilpin and Tuomas Sandholm.
		\newblock Better automated abstraction techniques for imperfect information
		games, with application to {T}exas {H}old'em poker.
		\newblock In {\em International Conference on Autonomous Agents and Multi-Agent
			Systems (AAMAS)}, pages 1168--1175, 2007.
		
		\bibitem{Gilpin08:Heads-up}
		Andrew Gilpin, Tuomas Sandholm, and Troels~Bjerre S{\o}rensen.
		\newblock A heads-up no-limit texas hold'em poker player: discretized betting
		models and automatically generated equilibrium-finding programs.
		\newblock In {\em Proceedings of the Seventh International Joint Conference on
			Autonomous Agents and Multiagent Systems-Volume 2}, pages 911--918.
		International Foundation for Autonomous Agents and Multiagent Systems, 2008.
		
		\bibitem{Jackson14:Time}
		Eric Jackson.
		\newblock A time and space efficient algorithm for approximately solving large
		imperfect information games.
		\newblock In {\em AAAI Workshop on Computer Poker and Imperfect Information},
		2014.
		
		\bibitem{Johanson13:Measuring}
		Michael Johanson.
		\newblock Measuring the size of large no-limit poker games.
		\newblock Technical report, University of Alberta, 2013.
		
		\bibitem{Johanson12:Finding}
		Michael Johanson, Nolan Bard, Neil Burch, and Michael Bowling.
		\newblock Finding optimal abstract strategies in extensive-form games.
		\newblock In {\em Proceedings of the Twenty-Sixth AAAI Conference on Artificial
			Intelligence}, pages 1371--1379. AAAI Press, 2012.
		
		\bibitem{Kroer17:Theoretical}
		Christian Kroer, Kevin Waugh, Fatma K{\i}l{\i}n\c{c}-Karzan, and Tuomas
		Sandholm.
		\newblock Theoretical and practical advances on smoothing for extensive-form
		games.
		\newblock In {\em Proceedings of the ACM Conference on Economics and
			Computation (EC)}, 2017.
		
		\bibitem{Lanctot09:Monte}
		Marc Lanctot, Kevin Waugh, Martin Zinkevich, and Michael Bowling.
		\newblock {M}onte {C}arlo sampling for regret minimization in extensive games.
		\newblock In {\em Proceedings of the Annual Conference on Neural Information
			Processing Systems (NIPS)}, pages 1078--1086, 2009.
		
		\bibitem{Littlestone94:Weighted}
		Nick Littlestone and M.~K. Warmuth.
		\newblock The weighted majority algorithm.
		\newblock {\em Information and Computation}, 108(2):212--261, 1994.
		
		\bibitem{Moravcik17:DeepStack}
		Matej Morav{\v c}{\'\i}k, Martin Schmid, Neil Burch, Viliam Lis{\'y}, Dustin
		Morrill, Nolan Bard, Trevor Davis, Kevin Waugh, Michael Johanson, and Michael
		Bowling.
		\newblock Deepstack: Expert-level artificial intelligence in heads-up no-limit
		poker.
		\newblock {\em Science}, 2017.
		
		\bibitem{Moravcik16:Refining}
		Matej Moravcik, Martin Schmid, Karel Ha, Milan Hladik, and Stephen Gaukrodger.
		\newblock Refining subgames in large imperfect information games.
		\newblock In {\em AAAI Conference on Artificial Intelligence (AAAI)}, 2016.
		
		\bibitem{Nash50:Eq}
		John Nash.
		\newblock Equilibrium points in n-person games.
		\newblock {\em Proceedings of the National Academy of Sciences}, 36:48--49,
		1950.
		
		\bibitem{Nesterov05:Excessive}
		Yurii Nesterov.
		\newblock Excessive gap technique in nonsmooth convex minimization.
		\newblock {\em SIAM Journal of Optimization}, 16(1):235--249, 2005.
		
		\bibitem{Sandholm10:State}
		Tuomas Sandholm.
		\newblock The state of solving large incomplete-information games, and
		application to poker.
		\newblock {\em AI Magazine}, pages 13--32, Winter 2010.
		\newblock Special issue on Algorithmic Game Theory.
		
		\bibitem{Sandholm15:Abstraction}
		Tuomas Sandholm.
		\newblock Abstraction for solving large incomplete-information games.
		\newblock In {\em AAAI Conference on Artificial Intelligence (AAAI)}, pages
		4127--4131, 2015.
		\newblock Senior Member Track.
		
		\bibitem{Sandholm15:Solving}
		Tuomas Sandholm.
		\newblock Solving imperfect-information games.
		\newblock {\em Science}, 347(6218):122--123, 2015.
		
		\bibitem{Schaeffer07:Checkers}
		Jonathan Schaeffer, Neil Burch, Yngvi Bj{\"o}rnsson, Akihiro Kishimoto, Martin
		M{\"u}ller, Robert Lake, Paul Lu, and Steve Sutphen.
		\newblock Checkers is solved.
		\newblock {\em Science}, 317(5844):1518--1522, 2007.
		
		\bibitem{Schnizlein09:Probabilistic}
		David Schnizlein, Michael Bowling, and Duane Szafron.
		\newblock Probabilistic state translation in extensive games with large action
		sets.
		\newblock In {\em Proceedings of the Twenty-First International Joint
			Conference on Artificial Intelligence}, pages 278--284, 2009.
		
		\bibitem{Silver16:Mastering}
		David Silver, Aja Huang, Chris~J Maddison, Arthur Guez, Laurent Sifre, George
		Van Den~Driessche, Julian Schrittwieser, Ioannis Antonoglou, Veda
		Panneershelvam, Marc Lanctot, et~al.
		\newblock Mastering the game of go with deep neural networks and tree search.
		\newblock {\em Nature}, 529(7587):484--489, 2016.
		
		\bibitem{Tammelin15:Solving}
		Oskari Tammelin, Neil Burch, Michael Johanson, and Michael Bowling.
		\newblock Solving heads-up limit texas hold'em.
		\newblock In {\em Proceedings of the International Joint Conference on
			Artificial Intelligence (IJCAI)}, pages 645--652, 2015.
		
		\bibitem{Waugh09:Strategy}
		Kevin Waugh, Nolan Bard, and Michael Bowling.
		\newblock Strategy grafting in extensive games.
		\newblock In {\em Proceedings of the Annual Conference on Neural Information
			Processing Systems (NIPS)}, 2009.
		
		\bibitem{Zinkevich07:Regret}
		Martin Zinkevich, Michael Johanson, Michael~H Bowling, and Carmelo Piccione.
		\newblock Regret minimization in games with incomplete information.
		\newblock In {\em Proceedings of the Annual Conference on Neural Information
			Processing Systems (NIPS)}, pages 1729--1736, 2007.
		
	\end{thebibliography}

	\clearpage
	
	\newpage
	
	\appendix
	
	\newpage
	
	\noindent {\LARGE \bf Appendix: Supplementary Material}
	
	\section{Maxmargin Solving}
	\label{sec:maxmargin}
	
	\emph{Maxmargin solving}~\cite{Moravcik16:Refining} is similar to Resolving, except that it seeks to improve $P_2$'s strategy in the subgame strategy as much as possible. While Resolving seeks a strategy for $P_2$ in $S$ that would simply dissuade $P_1$ from entering $S$, Maxmargin solving additionally seeks to punish $P_1$ as much as possible if $P_1$ nevertheless chooses to enter $S$. A \emph{subgame margin} is defined for each infoset in $S_{r}$, which represents the difference in value between entering the subgame versus choosing the alternative payoff. Specifically, for each infoset $I_1 \in S_{\textit{top}}$, the \emph{subgame margin} is
	\begin{equation}
	M^{\sigma^S}(I_1) = CBV^{\sigma_2}(I_1) - CBV^{\sigma^S_2}(I_1)
	\end{equation}
	
	In Maxmargin solving, a Nash equilibrium $\sigma^S$ for the augmented subgame described in Resolving subgame solving is computed such that the minimum margin over all $I_1 \in S_{\textit{top}}$ is maximized. Aside from maximizing the minimum margin, the augmented subgames used in Resolving and Maxmargin solving are identical.
	
	Given our base strategy in Coin Toss, Maxmargin solving would result in $P_2$ choosing Heads with probability $\frac{5}{8}$, Tails with probability $\frac{3}{8}$, and Forfeit with probability $0$.
	
	The augmented subgame can be solved in a way that maximizes the minimum margin by using a standard LP solver. In order to use iterative algorithms such as the Excessive Gap Technique~\cite{Nesterov05:Excessive,Gilpin12:First,Kroer17:Theoretical} or Counterfactual Regret Minimization (CFR)~\cite{Zinkevich07:Regret}, one can use the \emph{gadget game} described by Moravcik et al.~\cite{Moravcik16:Refining}. Details on the gadget game are provided in the Appendix. Our experiments used CFR.
	
	Maxmargin solving is safe. Furthermore, it guarantees that if every Player~1 best response reaches the subgame with positive probability through some infoset(s) that have positive margin, then exploitability is strictly lower than that of the blueprint strategy. While the theoretical guarantees are stronger, Maxmargin may lead to worse practical performance relative to Resolving when combined with the techniques discussed in Section~\ref{sec:error}, due to Maxmargin's greater tendency to overfit to assumptions in the model.
	
	\section{Description of Gadget Game}
	\label{sec:gadget}
	Solving the augmented subgame described in Maxmargin solving and Reach-Maxmargin solving will not, by itself, necessarily maximize the minimum margin. While LP solvers can easily handle this objective, the process is more difficult for iterative algorithms such as Counterfactual Regret Minimization (CFR) and the Excessive Gap Technique (EGT). For these iterative algorithms, the augmented subgame can be modified into a \emph{gadget game} that, when solved, will provide a Nash equilibrium to the augmented subgame and will also maximize the minimum margin~\cite{Moravcik16:Refining}. This gadget game is unnecessary when using distributional alternative payoffs, which is introduced in section~\ref{sec:distribution}.
	
	The gadget game differs from the augmented subgame in two ways. First, all $P_1$ payoffs that are reached from the initial infoset of $I_1 \in S_r$ are shifted by the alternative payoff of $I_1$, and there is longer an alternative payoff. Second, rather than the game starting with a chance node that determines $P_1$'s starting infoset, $P_1$ decides for herself which infoset to begin the game in. Specifically, the game begins with a $P_1$ node where each action in the node corresponds to an infoset $I_1$ in $S_r$. After $P_1$ chooses to enter an infoset $I_1$, chance chooses the precise node $h \in I_1$ in proportion to $\pi^{\sigma}_{-1}(h)$.
	
	By shifting all payoffs in the game by the size of the alternative payoff, the gadget game forces $P_1$ to focus on improving the performance of each infoset over some baseline, which is the goal of Maxmargin and Reach-Maxmargin solving. Moreover, by allowing $P_1$ to choose the infoset in which to enter the game, the gadget game forces $P_2$ to focus on maximizing the minimum margin.
	
\begin{figure}[!h]
		\centering
		\includegraphics[width=120mm]{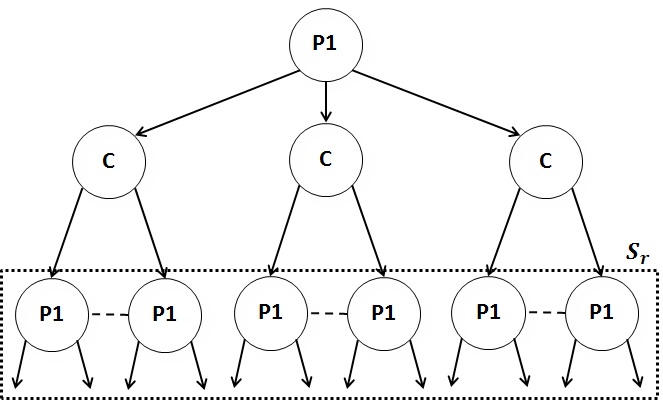}
		\caption{An example of a gadget game in Maxmargin refinement. $P_1$ picks the initial infoset she wishes to enter $S_r$ in. Chance then picks the particular node of the infoset, and play then proceeds identically to the augmented subgame, except all $P_1$ payoffs are shifted by the size of the alternative payoff and the alternative payoff is then removed from the augmented subgame.}
		\label{fig:gadget}
	\end{figure}

	Figure~\ref{fig:gadget} illustrates the gadget game used in Maxmargin and Reach-Maxmargin.
	
	\subsection{Distributional Alternative Payoffs}
	\label{sec:distribution}
	
	One problem with existing safe subgame-solving techniques is that they may ``overfit'' to the alternative payoffs, even when we use estimates. Consider for instance a subgame with two different $P_1$ infosets $I_1$ and $I'_1$ at the top. Assume $P_1$'s value for $I_1$ is estimated to be $1$, and for $I'_1$ is $10$. Now suppose during subgame solving, $P_2$ has a choice between two different strategies. The first sets $P_1$'s value in the subgame for $I_1$ to $0.99$ and for $I'_1$ to $9.99$. The second slightly increases $P_1$'s value for the subgame for $I_1$ to $1.01$ but dramatically lowers the value for $I'_1$ to $0$. The safe subgame-solving methods described so far would choose the first strategy, because the second strategy leaves one of the margins negative. However, intuitively, the second strategy is likely the better option, because it is more robust to errors in the model. For example, perhaps we are not confident that $10$ is the exact value, but instead believe its true value is normally distributed with $10$ as the mean and a standard deviation of $1$. In this case, we would prefer the strategy that lowers the value for $I'_1$ to $0$.
	
\begin{figure}[!h]
		\centering
		\includegraphics[width=120mm]{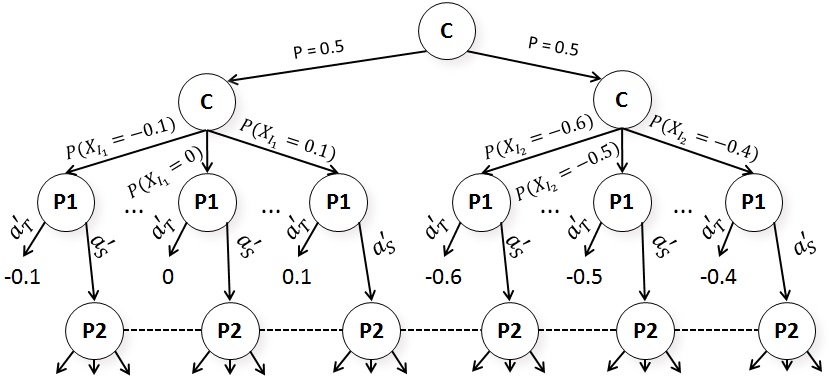}
		\caption{A visualization of the change in the augmented subgame from Figure~\ref{fig:resolve} when using distributional alternative payoffs.}
		\label{fig:distribution}
	\end{figure}

	To address this problem, we introduce a way to incorporate the modeling uncertainty into the game itself. Specifically, we introduce a new augmented subgame that makes subgame solving more robust to errors in the model. This augmented subgame changes the augmented subgame used in subgame Resolving (shown in Figure~\ref{fig:resolve}) so that the alternative payoffs are random variables, and $P_1$ is informed at the start of the augmented subgame of the values drawn from the random variables (but $P_2$ is not). The augmented subgame is otherwise identical. A visualization of this change is shown in Figure~\ref{fig:distribution}. As the distributions of the random variables narrow, the augmented subgame converges to the Resolve augmented subgame (but still maximizes the minimum margin when all margins are positive). As the distributions widen, $P_2$ seeks to maximize the sum over all margins, regardless of which are positive or negative.
	
	This modification makes the augmented subgame infinite in size because the random variables may be real-valued and $P_1$ could have a unique strategy for each outcome of the random variable. Fortunately, the special structure of the game allows us to arrive at a $P_2$ Nash equilibrium strategy for this infinite-sized augmented subgame by solving a much simpler gadget game.
	
	The gadget game is identical to the augmented subgame used in Resolve subgame solving (shown in Figure~\ref{fig:resolve}), except at each initial $P_1$ infoset $I_{r,1} \in S_r$, $P_1$ chooses action $a'_S$ (that is, chooses to enter the subgame rather than take the alternative payoff) with probability $P\big(X_{I_1} \le v(I_{r,1}, a'_S)\big)$, where $v(I_{r,1}, a'_S)$ is the expected value of action $a'_S$. (When solving via CFR, it is the expected value on each iteration, as described in CFR-BR~\cite{Johanson12:Finding}). This leads to Theorem~\ref{th:distributional}, which proves that solving this simplified gadget game produces a $P_2$ strategy that is a Nash equilibrium in the infinite-sized augmented subgame illustrated in Figure~\ref{fig:distribution}.
	
	\begin{theorem}
		\label{th:distributional}
		Let $S'$ be a Resolve augmented subgame and $S'_r$ its root. Let $S$ be a Distributional augmented subgame similar to $S'$, except at each infoset $I_{r,1} \in S_r$, $P_1$ observes the outcome of a random variable $X_{I_1}$ and the alternative payoff is equal to that outcome. If CFR is used to solve $S'$ except that the action leading to $S'$ is taken from each $I_{r,1} \in S'_r$ with probability $P\big(X_{I_1} \le v^t(I_{r,1}, a'_S)\big)$, where $v^t(I_{r,1},a'_S)$ is the value on iteration $t$ of action $a'_S$, then the resulting $P_2$ strategy $\sigma^{S'}_2$ in $S'$ is a $P_2$ Nash equilibrium strategy in $S$.
	\end{theorem}
	
	Another option which also solves the game but has better empirical performance relies on the \emph{softmax} (also known as \emph{Hedge}) algorithm~\cite{Littlestone94:Weighted}. This gadget game is more complicated, and is described in detail in Appendix~\ref{ap:hedge}. We use the softmax gadget game in our experiments.
	
	The correct distribution to use for the random variables ultimately depends on the actual unknown errors in the model. In our experiments for this technique, we set $X_{I_1} \sim \mathcal{N}\big(\mu_{I_1}, s^2_{I_1}\big)$, where $\mu_I$ is the blueprint value (plus any gifts). $s_{I_1}$ is set as the difference between the blueprint value of $I_1$, and the true (that is, unabstracted) counterfactual best response value of $I_1$. Our experiments show that this heuristic works well, and future research could yield even better options.
	
	\section{Hedge for Distributional Subgame Solving}
	\label{ap:hedge}
	In this paper we use CFR~\cite{Zinkevich07:Regret} with Hedge in $S_r$, which allows us to leverage a useful property of the
	Hedge algorithm~\cite{Littlestone94:Weighted} to update all the infosets resulting from outcomes of $X_{I_1}$ simultaneously.\footnote{Another option is to apply CFR-BR~\cite{Johanson12:Finding} only at the initial $P_1$ nodes when deciding between $a'_T$ and $a'_S$.} When using Hedge, action $a'_S$ in infoset $I_{r,1}$ in the augmented subgame is chosen on iteration $t$ with probability $\frac{e^{\eta_t \hat{v}(I_{r,1},a'_S)}}{e^{\eta_t \hat{v}(I_{r,1},a'_S)} + e^{\eta_t \hat{v}(I_{r,1},a'_T)}}$. Where $\hat{v}(I_{r,1},a'_T)$ is the observed expected value of action $a'_T$, $\hat{v}(I_{r,1},a'_S)$ is the observed expected value of action $a'_S$, and $\eta_t$ is a tuning parameter. Since, action $a'_S$ leads to identical play by both players for all outcomes of $X$, $\hat{v}(I_{r,1},a'_S)$ is identical for all outcomes of $X$. Moreover, $\hat{v}(I_{r,1},a'_T)$ is simply the outcome of $X_{I_1}$. So the probability that $a'_S$ is taken across all infosets on iteration $t$ is
	\begin{equation}
	\label{eq:reach}
	\int_{-\infty}^{\infty} \frac{e^{\eta_t \hat{v}(I_{r,1},a'_S)}}{e^{\eta_t \hat{v}(I_{r,1},a'_S)} + e^{\eta_t x}} f_{X_{I_1}}(x) dx
	\end{equation}
	where $f_{X_{I_1}}(x)$ is the pdf of $X_{I_1}$.
	In other words, if CFR is used to solve the augmented subgame, then the game being solved is identical to Figure~\ref{fig:resolve} except that action $a'_S$ is always chosen in infoset $I_1$ on iteration $t$ with probability given by (\ref{eq:reach}). In our experiments, we set the Hedge tuning parameter $\eta$ as suggested in~\cite{Brown17:Dynamic}: $\eta_t = \frac{\sqrt{\ln(|A(I_1)|)}}{3\sqrt{VAR(I_1)_t}\sqrt{t}}$, where $VAR(I_1)_t$ is the observed variance in the payoffs the infoset has received across all iterations up to $t$. In the subgame that follows $S_r$, we use CFR+ as the solving algorithm.
	
	\section{Scaling of Gifts}
	\label{sec:reachscale}
	
\begin{figure}[!h]
		\centering
		\includegraphics[width=120mm]{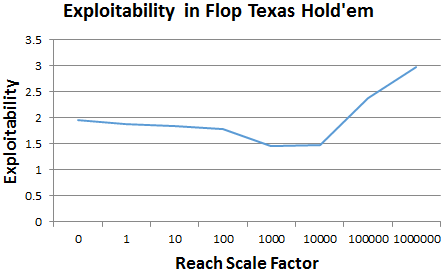}
		\caption{Exploitability in Flop Texas Hold'em of Reach-Maxmargin as we scale up the size of gifts.}
		\label{fig:reachscale}
	\end{figure}

	To retain the theoretical guarantees of Reach subgame solving, one must ensure that the gifts assigned to reachable subgames do not (in aggregate) exceed the original gift. That is, if $g(I_1)$ is a gift at infoset $I_1$, we must ensure that $CBV^{\sigma^*_2}(I_1) \le CBV^{\sigma_2}(I_1) + g(I_1)$. In this paper we accomplish this by increasing the margin of an infoset $I'_1$, where $I_1 \sqsubseteq I'_1$, by at most $g(I_1)$. However, empirical performance may improve if the increase to margins due to gifts is scaled up by some factor. In most games we experimented on, exploitability decreased the further the gifts were scaled. However, Figure~\ref{fig:reachscale} shows one case in which we observe the exploitability increasing when the gifts are scaled up too far. The graph shows exploitability when the gifts are scaled by various factors. At 0, the algorithm is identical to Maxmargin. at 1, the algorithm is the theoretically correct form of Reach-Maxmargin. Optimal performance in this game occurs when the gifts are scaled by a factor of about $1,000$. Scaling the gifts by $100,000$ leads to performance that is worse than Maxmargin subgame solving. This empirically demonstrates that while scaling up gifts may lead to better performance in some cases (because an entire gift is unlikely to be used in every subgame that receives one), it may also lead to far worse performance in some cases.
	
	\section{Rules for Poker Variants}
	\label{sec:rules}
	Our experiments are conducted on heads-up no-limit Texas hold'em (HUNL), as well as smaller-scale variants we call no-limit flop hold'em (NLFH) and no-limit turn hold'em (NLTH). We begin by describing the rules of HUNL.
	
	In the form of HUNL discussed in this paper, each player starts a hand with \$20,000. One player is designated $P_1$, while the other is $P_2$. This assignment alternates between hands. HUNL consists of four rounds of betting. On a round of betting, each player can choose to either fold, call, or raise. If a player folds, that player immediately surrenders the pot to the opponent and the game ends. If a player calls, that players places a number of chips in the pot equal to the opponent's contribution. If a player raises, that player adds more chips to the pot than the opponent's contribution. A round of betting ends after a player calls. Players can continue to go back and forth with raises in a round until one of them runs out of chips.
	
	If either player chooses to raise first in a round, they must raise a minimum of \$100. If a player raises after another player has raised, that raise must be greater than or equal to the last raise. The maximum amount for a bet or raise is the remainder of that player's chip stack, which in our model is \$20,000 at the beginning of a game.
	
	At the start of HUNL, both players receive two private cards from a standard 52-card deck. $P_1$ must place a \emph{big blind} of \$100 in the pot, while $P_2$ must place a \emph{small blind} of \$50 in the pot. There is then a round of betting (the \emph{preflop}), starting with $P_2$. When the round ends, three \emph{community} cards are dealt face up between the players. There is then another round of betting (the \emph{flop}), starting with $P_1$ this time. After the round of betting completes, another community card is dealt face up, and another round of betting commences starting with $P_1$ (the \emph{turn}). Finally, one more community card is dealt face up, and a final betting round occurs (the \emph{river}), again starting with $P_1$. If neither player folds before the final betting round completes, the player with the best five-card poker hand, constructed from their two private cards and the five face-up community cards, wins the pot. In the case of a tie, the pot is split evenly.
	
	NLTH is similar to no-limit Texas hold'em except there are only three rounds of betting (the preflop, flop, and turn) in which there are two options for bet sizes. There are also only four community cards. NLFH is similar except there are only two rounds of betting (the preflop and flop), and three community cards.
	
	We experiment with two versions of NLFH, one small and one large, which include only a few of the available actions in each infoset. The small game requires $1.1$ GB to store the unabstracted strategy as double-precision floats. The large game requires $4$ GB. NLTH requires $35$ GB to store the unabstracted strategy.
	
	\section{Proof of Theorem~\ref{th:multiple}}
	
	\begin{proof}
		Assume $M_{r}^{\sigma^S}(I_1) \ge 0$ for every infoset $I_1$ and assume $\pi_1^{BR(\sigma'_2)}(I^*_1) > 0$ for some $I^*_1 \in S_{\textit{top}}$ and let $\epsilon = M_{r}(I^*_1)$. Define $\pi^{\sigma}_{-1}(I_1) = \sum_{h \in I_1} \pi^{\sigma}_{-1}(h)$ and define $\pi^{\sigma}_{-1}(I_1, I'_1) = \sum_{h \in I_1, h' \in I'_1} \pi^{\sigma}_{-1}(h,h')$.
		
		We show that for every $P_1$ infoset $I_1 \sqsubseteq I^*_1$ where $P(I_1) = P_1$, 
		\begin{multline}
		CBV^{\sigma'_2}(I_1) \le CBV^{\sigma^{-S}_2}(I_1) + \\ \sum_{I''_1 \cdot a'' \sqsubseteq I_1 \mid P(I''_1) = P_1} \big(\lfloor CBV^{\sigma^{-S}_2}(I''_1) - CBV^{\sigma^{-S}_2}(I''_1,a'') \rfloor\big) - \sum_{h \in I_1, h^* \in I^*_1} \pi^{\sigma_2}_{-1}(h, h^*)\epsilon
		\label{eq:reach_goal}
		\end{multline}
		By the definition of $M^{\sigma^S}_r(I^*_1)$ this holds for $I^*_1$ itself. Moreover, the condition holds for every other $I_1 \in S_{\textit{top}}$, because by assumption every margin is nonnegative and $\pi_{-1}^{\sigma_2}(I_1,I^*_1) = 0$ for any $I_1 \in S_{\textit{top}}$ where $I_1 \ne I^*_1$. The condition also clearly holds for any $I_1$ with no descendants in $S$ because then $\pi_{-1}^{\sigma_2}(I_1,I^*_1) = 0$ and $\sigma'_2(h) = \sigma^{-S}_2(h)$ in all $P_2$ nodes following $I_1$. This satisfies the base step. We now move on to the inductive step.
		
		Let $Succ(I_1, a)$ be the set of earliest-reachable $P_1$ infosets following $I_1$ such that $P(I'_1) = P_1$ for $I' \in Succ(I_1,a)$. Formally, $I'_1 \in Succ(I_1, a)$ if $P(I'_1) = P_1$ and $I_1 \cdot a \sqsubseteq I'_1$ and for any other $I''_1 \in Succ(I_1,a)$, $I''_1 \not \sqsubset I'_1$. Then 
		\begin{multline}
		CBV^{\sigma'_2}(I_1,a) = CBV^{\sigma^{-S}_2}(I_1,a)+ \\ \sum_{I'_1 \in Succ(I_1,a)} \pi_{-1}^{\sigma'_2}(I_1,I'_1)(CBV^{\sigma_2'}(I'_1) - CBV^{\sigma^{-S}_2}(I'_1))
		\end{multline}
		Assume that every $I'_1 \in Succ(I_1,a)$ satisfies (\ref{eq:reach_goal}). Then	
		\begin{multline}
		CBV^{\sigma'_2}(I_1,a) \le CBV^{\sigma^{-S}_2}(I_1,a) - \pi_{-1}^{\sigma_2}(I_1,I^*_1)\epsilon + \\\sum_{I'_1 \in Succ(I_1,a)} \pi_{-1}^{\sigma_2}(I_1,I'_1) \Big(\sum_{I''_1 \cdot a'' \sqsubseteq I'_1 \mid P(I''_1) = P_1}\big(\lfloor CBV^{\sigma^{-S}_2}(I''_1) - CBV^{\sigma^{-S}_2}(I''_1,a'')\rfloor\big) \Big)
		\nonumber
		\end{multline}
		\begin{multline}
		CBV^{\sigma'_2}(I_1,a) \le CBV^{\sigma^{-S}_2}(I_1) - \Big(CBV^{\sigma^{-S}_2}(I_1) - CBV^{\sigma^{-S}_2}(I_1,a) \Big)-  \pi_{-1}^{\sigma_2}(I_1,I^*_1)\epsilon + \\\sum_{I'_1 \in Succ(I_1,a)} \pi_{-1}^{\sigma_2}(I_1,I'_1) \Big(\sum_{I''_1 \cdot a'' \sqsubseteq I'_1 \mid P(I''_1) = P_1}\big(\lfloor CBV^{\sigma^{-S}_2}(I''_1) - CBV^{\sigma^{-S}_2}(I''_1,a'')\rfloor\big) \Big)
		\nonumber
		\end{multline}
		Since $\lfloor CBV^{\sigma^{-S}_2}(I_1) - CBV^{\sigma^{-S}_2}(I_1,a)\rfloor \le CBV^{\sigma^{-S}_2}(I_1) - CBV^{\sigma^{-S}_2}(I_1,a_1)$ so we get
		\begin{multline}
		CBV^{\sigma'_2}(I_1,a) \le CBV^{\sigma^{-S}_2}(I_1) - \lfloor (CBV^{\sigma^{-S}_2}(I_1) - CBV^{\sigma^{-S}_2}(I_1,a) \rfloor -  \pi_{-1}^{\sigma_2}(I_1,I^*_1)\epsilon + \\\sum_{I'_1 \in Succ(I_1,a)} \pi_{-1}^{\sigma_2}(I_1,I'_1) \Big(\sum_{I''_1 \cdot a'' \sqsubseteq I'_1 \mid P(I''_1) = P_1}\big(\lfloor CBV^{\sigma^{-S}_2}(I''_1) - CBV^{\sigma^{-S}_2}(I''_1,a'')\rfloor\big) \Big)
		\nonumber
		\end{multline}
		\begin{multline}
		CBV^{\sigma'_2}(I_1,a) \le CBV^{\sigma^{-S}_2}(I_1) - \pi_{-1}^{\sigma_2}(I_1,I^*_1)\epsilon + \\\sum_{I'_1 \in Succ(I_1,a)} \pi_{-1}^{\sigma_2}(I_1,I'_1) \Big(\sum_{I''_1 \cdot a'' \sqsubseteq I_1 \mid P(I''_1) = P_1}\big(\lfloor CBV^{\sigma^{-S}_2}(I''_1) - CBV^{\sigma^{-S}_2}(I''_1,a'')\rfloor\big) \Big)
		\nonumber
		\end{multline}
		\begin{equation}
		CBV^{\sigma'_2}(I_1,a_1) \le CBV^{\sigma^{-S}_2}(I_1) - \pi_{-1}^{\sigma_2}(I_1,I^*_1)\epsilon + \sum_{I''_1 \cdot a'' \sqsubseteq I_1 \mid P(I''_1) = P_1}\big(\lfloor CBV^{\sigma^{-S}_2}(I''_1) - CBV^{\sigma^{-S}_2}(I''_1,a''_1)\rfloor\big)
		\nonumber
		\end{equation}
		Since $\pi_1^{BR(\sigma'_2)}(I^*_1) > 0$, and action $a$ leads to $I^*_1$, so by definition of a best response, $CBV^{\sigma'_2}(I_1, a) = CBV^{\sigma'_2}(I_1)$. Thus,
		\begin{equation}
		CBV^{\sigma'_2}(I_1) \le CBV^{\sigma^{-S}_2}(I_1) - \pi_{-1}^{\sigma_2}(I_1,I^*_1)\epsilon + \sum_{I''_1 \cdot a'' \sqsubseteq I_1 \mid P(I''_1) = P_1}\big(\lfloor CBV^{\sigma^{-S}_2}(I''_1) - CBV^{\sigma^{-S}_2}(I''_1,a'')\rfloor\big)
		\nonumber
		\end{equation}
		which satisfies the inductive step.
		
		Applying this reasoning to the root of the entire game, we arrive at $exp(\sigma_2') \le exp(\sigma^{-S}_2) - \pi^{\sigma_2}_{-1}(I^*_1)\epsilon$.
	\end{proof}
	
\section*{Proof of Theorem~\ref{th:estimate}}
\begin{proof}

Without loss of generality, we assume that it is player $P_2$ who conducts subgame solving. 
We define a node $h$ in a subgame $S$ as \emph{earliest-reachable} if there does not exist a node $h' \in S$ such that $h' \prec h$. For each earliest-reachable node $h \in S$, let $h_r$ be its parent and $a_S$ be the action leading to $h$ such that $h_r \cdot a_S = h$. We require $h_r$ to be a $P_1$ node; if it is not, then we can simply insert a $P_1$ node with only a single action between $h_r$ and $h$. Let $S_r$ be the set of all $h_r$ for $S$.

Applying subgame solving to subgames as they are reached during play is equivalent to applying subgame solving to every subgame before play begins, so we can phrase what follows in the context of all subgames being solved before play begins. Let $\sigma'_2$ be the $P_2$ strategy produced after subgame solving is applied to every subgame. We show inductively that for any $P_1$ infoset $I_1 \not \in \mathcal{S}$ where it is $P_1$'s turn to move (i.e., $P(I_1)=P_1$), the counterfactual best response values for $P_1$ satisfy
\begin{equation}
\label{eq:maxmargin_inductive}
CBV^{\sigma'_2}(I_1) \le CBV^{\sigma^*_2}(I_1) + 2 \Delta
\end{equation}
Define $\textit{Succ}(I_1, a)$ as the set of infosets belonging to $P_1$ that follow action $a$ in $I_1$ and where it is $P_1$'s turn and where $P_1$ has not had a turn since $a$, as well as terminal nodes follow action $a$ in $I_1$ without $P_1$ getting a turn.
Formally, a terminal node $z \in Z$ is in $\textit{Succ}(I_1,a)$ if there exists a history $h \in I_1$ such that $h \cdot a \preceq z$ and there does not exist a history $h'$ such that $P(h') = P_1$ and $h \cdot a \preceq h' \prec z$. Additionally, an infoset $I'_1$ belonging to $P_1$ is in $\textit{Succ}(I_1, a)$ if $P(I'_1) = P_1$ and $I_1 \cdot a \preceq I'_1$ and there does not exist an earlier infoset $I''_1$ belonging to $P_1$ such that $P(I''_1) = P_1$ and $I' \cdot a \preceq I''_1 \prec I'_1$. Define $\textit{Succ}(I_1)$ as $\cup_{a \in A(I_1)} \textit{Succ}(I_1,a)$. Similarly, we define $\textit{Succ}(h,a)$ as the set of histories belonging to $P(h)$, or terminals, that follow action $a$ and where $P(h)$ has not had a turn since $a$. Formally, $h' \in \textit{Succ}(h,a)$ if either $P(h') = P(h)$ or $P(h') \in Z$ and $h \cdot a \preceq h'$ and there does not exist a history $h''$ such that $P(h'') = P(h)$ and $h \cdot a \preceq h'' \prec h'$.

Now we define a level $L$ for each $P_1$ infoset where it is $P_1$'s turn and the infoset is not in the set of subgames $\mathcal{S}$. 
\begin{itemize}
	\item For immediate parents of subgames we define the level to be zero: for all $I_1 \in S_r$ for any subgame $S \in \mathcal{S}$,  $L(I_1) = 0$.
	\item For infoset that are not ancestors of subgames, we define the level to be zero: $L(I_1) = 0$ for any infoset $I_1$ that is not an ancestor of a subgame in $\mathcal{S}$.
	\item For all other infosets, the level is one greater than the greatest level of its successors: $L(I_1) = \ell + 1$ where $\ell = \max_{I'_1 \in \textit{Succ}(I_1)}L(I'_1)$ where $L(z) = 0$ for terminal nodes $z$.
\end{itemize}

\subsection*{Base case of induction}
First consider infosets $I_1 \in S_r$ for some subgame $S \in \mathcal{S}$. 
We define $M^{\sigma'_2}(I_1) = v^{\sigma}(I_1,a_S) - CBV^{\sigma'_2}(I_1,a_S)$. Consider a subgame $S \in \mathcal{S}$. Estimated-Maxmargin subgame solving arrives at a strategy $\sigma'_2$ such that $\min_{I_1 \in S_r}M^{\sigma'_2}(I_1)$ is maximized. By the assumption in the theorem statement, $|v^{\sigma}(I_1,a_S) - CBV^{\sigma^*_2}(I_1,a_S)| \le \Delta$ for all $I_1 \in S_r$. Thus, $\sigma^*_2$ satisfies $\min_{I_1 \in S_r}M^{\sigma^*_2}(I_1) \ge -\Delta$ and therefore $\min_{I_1 \in S_r}M^{\sigma'_2}(I_1) \ge -\Delta$, because Estimated-Maxmargin subgame solving could, at least, arrive at $\sigma'_2 = \sigma^*_2$. From the definition of $M^{\sigma'_2}(I_1)$, this implies that for all $I_1 \in S_r$, $CBV^{\sigma'_2}(I_1,a_S) \le v^{\sigma}(I_1,a_S) + \Delta$. Since by assumption $v^{\sigma}(I_1,a_S) \le CBV^{\sigma^*_2}(I_1,a_S) + \Delta$, this gives us $CBV^{\sigma'_2}(I_1,a_S) \le CBV^{\sigma^*_2}(I_1,a_S) + 2\Delta$.

Now consider infosets $I_1$ that are not ancestors of any subgame in $\mathcal{S}$. By definition, for all $h$ such that $h \succeq I_1$ or $I_1 \succeq h$, and $P(h) = P_2$, $\sigma^*_2(I_2(h)) = \sigma_2(I_2(h)) = \sigma'_2(I_2(h))$. Therefore, $CBV^{\sigma'_2}(I_1) = CBV^{\sigma^*_2}(I_1)$. 

So, we have shown that (\ref{eq:maxmargin_inductive}) holds for any $I_1$ such that $L(I_1) = 0$.

\subsection*{Inductive step}
Now assume that (\ref{eq:maxmargin_inductive}) holds for any $P_1$ infoset $I_1$ where $P(I_1) = P_1$ and $I_1 \not \in \mathcal{S}$ and $L(I_1) \le \ell$. Consider an $I_1$ such that $P(I_1) = P_1$ and $I_1 \not \in \mathcal{S}$ and $L(I_1) = \ell + 1$.

From the definition of $CBV^{\sigma'_2}(I_1,a)$, we have that for any action $a \in A(I_1)$,
\begin{equation}
CBV^{\sigma'_2}(I_1, a) = \Big(\sum_{h \in I_1} \Big(\big(\pi_{-1}^{\sigma'_2}(h)\big) \big(v^{\langle CBR(\sigma'_2), \sigma'_2\rangle}(h \cdot a)\big)\Big)\Big) / \sum_{h \in I_1} \pi_{-1}^{\sigma'_2}(h)
\end{equation}
Since for any $h \in I_1$ there is no $P_1$ action between $a$ and reaching any $h' \in \textit{Succ}(h,a)$, so $\pi_1^{\sigma'_2}(h \cdot a, h') = 1$. Thus,
\begin{equation}
CBV^{\sigma'_2}(I_1, a) = \Big(\sum_{h \in I_1} \Big(\pi_{-1}^{\sigma'_2}(h) \sum_{h' \in \textit{Succ}(h, a)} \pi^{\sigma'_2}_{-1}(h, h') \big(v^{\langle CBR(\sigma'_2), \sigma'_2\rangle}(h')\big)\Big)\Big) / \sum_{h \in I_1} \pi_{-1}^{\sigma'_2}(h)
\end{equation}
\begin{equation}
CBV^{\sigma'_2}(I_1, a) = \Big(\sum_{h \in I_1} \sum_{h' \in \textit{Succ}(h, a)} \Big(\big(\pi^{\sigma'_2}_{-1}(h')\big) v^{\langle CBR(\sigma'_2), \sigma'_2\rangle}(h')\Big)\Big) / \sum_{h \in I_1} \pi_{-1}^{\sigma'_2}(h)
\end{equation}
Since the game is perfect recall, $\sum_{h \in I_1} \sum_{h' \in \textit{Succ}(h,a)} f(h') = \sum_{I'_1 \in \textit{Succ}(I_1,a)} \sum_{h' \in I'_1} f(h')$ for any function $f$. Thus,
\begin{equation}
CBV^{\sigma'_2}(I_1, a) = \Big(\sum_{I'_1 \in \textit{Succ}(I_1,a)} \sum_{h' \in I'_1} \Big(\big(\pi_{-1}^{\sigma'_2}(h')\big)\big(v^{\langle CBR(\sigma'_2), \sigma'_2\rangle}(h')\big)\Big)\Big) / \sum_{h \in I_1} \pi_{-1}^{\sigma'_2}(h)
\end{equation}
From the definition of $CBV^{\sigma'_2}(I'_1)$ we get
\begin{equation}
CBV^{\sigma'_2}(I_1, a) = \Big(\sum_{I'_1 \in \textit{Succ}(I_1,a)} \big(CBV^{\sigma'_2}(I'_1) \sum_{h' \in I'_1} \pi_{-1}^{\sigma'_2}(h')\big)\Big) / \sum_{h \in I_1} \pi_{-1}^{\sigma'_2}(h)
\label{eq:cbv}
\end{equation}
Since (\ref{eq:maxmargin_inductive}) holds for all $I'_1 \in \textit{Succ}(I_1,a)$, so
\begin{equation}
CBV^{\sigma'_2}(I_1, a) \le \Big(\sum_{I'_1 \in \textit{Succ}(I_1,a)} \big((CBV^{\sigma^*_2}(I'_1) + 2\Delta) \sum_{h' \in I'_1} \pi_{-1}^{\sigma'_2}(h')\big)\Big) / \sum_{h \in I_1} \pi_{-1}^{\sigma'_2}(h)
\end{equation}
Since $P_2$'s strategy is fixed according to $\sigma_2$ outside of $\mathcal{S}$, so for all $I_1 \not \in \mathcal{S}$, $\pi_{-1}^{\sigma'}(I_1) = \pi_{-1}^{\sigma}(I_1) = \pi_{-1}^{\sigma^*}(I_1)$. Therefore,
\begin{equation}
CBV^{\sigma'_2}(I_1, a) \le \Big(\sum_{I'_1 \in \textit{Succ}(I_1,a)} \big((CBV^{\sigma^*_2}(I'_1) + 2\Delta) \sum_{h' \in I'_1} \pi_{-1}^{\sigma^*_2}(h')\big)\Big) / \sum_{h \in I_1} \pi_{-1}^{\sigma^*_2}(h)
\end{equation}
Pulling out the $2 \Delta$ constant and applying equation (\ref{eq:cbv}) for $CBV^{\sigma^*_2}(I_1,a)$ we get
\begin{equation}
CBV^{\sigma'_2}(I_1, a) \le CBV^{\sigma^*}(I_1,a) + 2\Delta \Big(\big(\sum_{I'_1 \in \textit{Succ}(I_1,a)} \sum_{h' \in I'_1} \pi_{-1}^{\sigma^*_2}(h')\big) / \sum_{h \in I_1} \pi_{-1}^{\sigma^*_2}(h) \Big)
\end{equation}
Since $\big(\sum_{I'_1 \in \textit{Succ}(I_1,a)} \sum_{h' \in I'_1} \pi_{-1}^{\sigma^*_2}(h')\big) = \sum_{h \in I_1} \pi_{-1}^{\sigma^*_2}(h)$ we arrive at
\begin{equation}
CBV^{\sigma'_2}(I_1, a) \le CBV^{\sigma^*}(I_1,a) + 2\Delta
\end{equation}

Thus, (\ref{eq:maxmargin_inductive}) holds for $I_1$ as well and the inductive step is satisfied. Extending (\ref{eq:maxmargin_inductive}) to the root of the game, we see that $\textit{exp}(\sigma'_2) \le \textit{exp}(\sigma^*_2) + 2 \Delta$.
\end{proof}
	
	\section{Proof of Theorem~\ref{th:distributional}}
	\begin{proof}
		We prove inductively that using CFR in $S'$ while choosing the action leading to $S'$ from each $I_1 \in S'_r$ with probability $P\big(X_{I_1} \le v^t(I_1, a'_S)\big)$ results in play that is identical to CFR in $S$ and CFR-BR~\cite{Johanson12:Finding} in $S_r$, which converges to a Nash equilibrium.
		
		For each $P_2$ infoset $I'_2$ in $S'$ where $P(I'_2) = P_2$, there is exactly one corresponding infoset $I_2$ in $S$ that is reached via the same actions, ignoring random variables. Each $P_1$ infoset $I'_1$ in $S'$ where $P(I'_1) = P_1$ corresponds to a set of infosets in $S$ that are reached via the same actions, where the elements in the set differ only by the outcome of the random variables. We prove that on each iteration, the instantaneous regret for these corresponding infosets is identical (and therefore the average strategy played in the $P_2$ infosets over all iterations is identical).
		
		At the start of the first iteration of CFR, all regrets are zero. Therefore, the base case is trivially true. Now assume that on iteration $t$, regrets are identical for all corresponding infosets. Then the strategies played on iteration $t$ in $S$ are identical as well.
		
		First, consider an infoset $I'_1$ in $S'$ and a corresponding infoset $I_1$ in $S$. Since the remaining structure of the game is identical beyond $I'_1$ and $I_1$, and because $P_2$'s strategies are identical in all $P_2$ infosets encountered, so the immediate regret for $I'_1$ and $I_1$ is identical as well.
		
		Next, consider a $P_1$ infoset $I_{1,x}$ in $S_r$ in which the random variable $X_{I_1}$ has an observed value of $x$. Let the corresponding $P_1$ infoset in $S'_r$ be $I'_1$. Since CFR-BR is played in this infoset, and since action $a'_T$ leads to a payoff of $x$, so $P_1$ will choose action $a'_S$ with probability 1 if $x \ge a'_T$ and with probability 0 otherwise. Thus, for all infosets in $S_r$ corresponding to $I'_1$, action $a'_S$ is chosen with probability $P\big(X_{I_1} \le v(I_1, a'_S)\big)$.
		
		Finally, consider a $P_2$ infoset $I_2$ in $S$ and its corresponding infoset $I'_2$ in $S'$. Since in both cases action $a'_T$ is taken in $S_r$ with probability $P\big(X_{I_1} \le v(I_1, a'_S)\big)$, and because $P_1$ plays identically between corresponding infosets in $S$ and $S'$, and because the structure of the game is otherwise identical, so the immediate regret for $I'_1$ and $I_1$ is identical as well.
	\end{proof}
	
\end{document}